\documentclass[sn-mathphys.bst,iicol]{sn-jnl}

\usepackage{graphicx}%
\usepackage{multirow}%
\usepackage{amsmath,amssymb,amsfonts}%
\usepackage{amsthm}%
\usepackage{mathrsfs}%
\usepackage[title]{appendix}%
\usepackage{xcolor}%
\usepackage{textcomp}%
\usepackage{manyfoot}%
\usepackage{booktabs}%
\usepackage{algorithm}%
\usepackage{algorithmicx}%
\usepackage{algpseudocode}%
\usepackage{listings}%
\usepackage{subcaption}

\usepackage{lipsum}  

\makeatletter
\renewcommand\paragraph{\@startsection{paragraph}{4}{\z@}%
            {-3.25ex\@plus -1ex \@minus -.2ex}%
            {1.5ex \@plus .2ex}%
            {\normalfont\normalsize\bfseries}}
\makeatother

\renewcommand{\theparagraph}{\thesubsubsection.\arabic{paragraph}}

\theoremstyle{thmstyleone}%
\theoremstyle{thmstyletwo}%

\theoremstyle{thmstylethree}%

\usepackage{float}
\usepackage{array}  
\usepackage{booktabs} 
\usepackage{multirow}
\usepackage{makecell} 
\usepackage{eurosym}

\usepackage{cite}

\usepackage{geometry}
\usepackage{graphics}
\usepackage{caption}
\newcommand{\mybluehl}[1]{\textcolor{black}{#1}} 

\usepackage{fancyhdr}

\begin{document}

\title[Tactile interaction with social robots influences attitudes and behaviour]{Tactile interaction with social robots influences attitudes and behaviour}

\author*{\fnm{Qiaoqiao} \sur{Ren}}\email{Qiaoqiao.Ren@ugent.be}

\author{\fnm{Tony} \sur{Belpaeme}}\email{Tony.Belpaeme@ugent.be}

\affil{\orgdiv{AIRO - IDLab}, \orgname{Ghent University - imec}, \orgaddress{\street{Technologiepark 126}, \city{Gent}, \postcode{9052}, \country{Belgium}}}


\abstract{
Tactile interaction plays an essential role in human-to-human interaction. People gain comfort and support from tactile interactions with others and touch is an important predictor for trust. While touch has been explored as a communicative modality in HCI and HRI, we here report on two studies in which touching a social robot is used to regulate people's stress levels and consequently their actions.

In the first study, we look at whether different intensities of tactile interaction result in a physiological response related to stress, and whether the interaction impacts risk-taking behaviour and trust. We let 38 participants complete a Balloon Analogue Risk Task (BART), a computer-based game that serves as a proxy for risk-taking behaviour. In our study, participants are supported by a robot during the BART task. The robot builds trust and encourages participants to take more risk. The results show that affective tactile interaction with the robot increases participants' risk-taking behaviour, but gentle affective tactile interaction increases comfort and lowers stress whereas high-intensity touch does not. We also find that male participants exhibit more risk-taking behaviour than females while being less stressed. 
 Based on this experiment, a second study is used to ascertain whether these effects are caused by the social nature of tactile interaction or by the physical interaction alone. For this, instead of a social robot, participants now have a tactile interaction with a non-social device. 
 The non-social interaction does not result in any effect, leading us to conclude that tactile interaction with humanoid robots is a social phenomenon rather than a mere physical phenomenon.}

\keywords{Affective touch; human-robot touch; tactile interaction; haptic interaction; nonverbal communication; peer pressure; risk-taking behaviour; heart rate variability.}



\maketitle

\thispagestyle{empty}
\pagestyle{fancy}
\renewcommand{\headrulewidth}{0pt}

\lhead{}
\chead{}
\rhead{}
\lfoot{}
\cfoot{\thepage}
\rfoot{}

\section{Introduction}\label{sec1}

The application of haptic and tactile interaction is expected to see significant growth in human-robot interaction (HRI) research. Haptic interaction, in this context a physical interaction with digital media \cite{lee2006physically}, conveys emotions and promotes a sense of interpersonal closeness \cite{andreasson2018affective}. As robots enter social spaces, they will have the opportunity to join in similar physical interactions. And as tactile interaction has been found to support communication and strengthen connections between individuals, often contributing to overall well-being \cite{zhou2021tactile,chan2020effect}, we expect similar effects to manifest in people when engaging in tactile interactions with robots.

\subsection{Physical interaction between people and robots}

People often bond with friends and team members by shaking hands, giving high fives, or bumping fists. As robots start to become a part of our social lives, they are likely to take part in these kinds of physical greetings and interactions \cite{fitter2016equipping}. Physical interaction with robots is relatively uncommon for several reasons. One significant issue is the lack of safety during haptic interactions \cite{li2017touching}. Simple greetings like hugging, holding, or shaking hands already form a safety concern. Although many robots are explicitly designed for human-robot interaction (HRI), tactile or haptic interaction has in many cases been an afterthought. For example, the two best-selling social robots, the United Robotics Group's Nao and Pepper robots, both have pinch points that can trap fingers making these robots unsuitable for tactile interaction. Only a limited number of robots are explicitly designed to allow physical human-robot interaction, with most other robots prone to suffer breakage when being manipulated or handled with forces that would not be uncommon in interactions between people \cite{alami2006safe}. 


There is a significant body of literature that reports on the impact of physical and haptic interaction between people on attitudes, emotions, and performance. For example, even subtle physical contact between individuals can enhance feelings of security and increase risk-taking behaviours \cite{hri1}. Similarly, haptic and physical interaction has been shown to alter physiological responses in human-robot interaction. The current study aims to investigate the impact of the \emph{intensity} of tactile interaction on people's attitudes, stress levels, arousal, and task performance with a robot, with a specific focus on understanding gender differences in physical and physiological responses during a task. Ultimately, the aim is to contribute to an understanding of the potential benefits of human-robot tactile interaction for physical and physiological well-being.

Affective touch has been shown to reduce feelings of stress and anxiety. Affective touch increases dopamine and serotonin levels, leading to improvements in mood and reductions in discomfort \cite{von2017soothing}. Recent research has also suggested that affective touch can be an effective means of reducing social exclusion, with studies finding that participants are more willing to participate in experimental tasks when touching is perceived to be in line with their goals \cite{sailer2022meaning}. Affective touch also has the potential to reduce social pain associated with exclusion \cite{von2017soothing}. Studies have found that slow, emotive touch influences the perception of physical pain and appears to be mediated by a distinct tactile neurophysiological system \cite{hri4}.  It is believed that the interpretation of affective touch is determined by the perceived meaning of the touch, which is influenced by the match between the touch's purpose and the receiver's goals \cite{jakubiak2016keep}. 
 The literature suggests that social presence, such as human tactile interaction, can increase human trust and touch may be instrumental for building trust \cite{walter2013trust}.
While touch between people who are unfamiliar with each other is often short, under certain circumstances prolonged touch between parties that are not intimate can be comfortable, with massage being a prime example here \cite{sailer2022meaning}. Next to being merely relaxing, touch also conveys emotion and meaning in a larger social context
\cite{jakubiak2016keep}.

There is a small but significant body of literature that describes the use of robots to support tactile communication with people, and research has shown that touch can influence emotions, stress levels, and physiological arousal \cite{ren2023behavioural, ren2022tactile}. For instance, it was found that tactile interactions with a PARO robot can decrease pain perception and elevate salivary oxytocin levels \cite{hri5}. Li \emph{et al.} found that touching a humanoid robot with identifiable body parts leads to significant changes in arousal levels, providing evidence that tactile HRI elicits physiological responses \cite{li2017touching}. Chan \emph{et al.} demonstrated that remotely delivered affective touch via a robot leads to changes in psychological experience and skin conductance response (SCR) \cite{chan2020effect}. 

\subsection{Risk-taking behaviour}

This paper focuses on the impact of physical and haptic interaction on decision taking, and specifically on \emph{risk-taking behaviour}. Risk-taking is a fundamental human activity with important implications for individuals' health and social lives. One significant factor that has been shown to influence risk-taking behaviour is \emph{peer pressure}  \cite{sandstrom2011power}. When individuals observe others, particularly friends or peers, engaging in risky behaviour and seeking validation, they are more likely to participate in those behaviours themselves. This is evident in various situations, such as in drivers who engage in reckless driving when encouraged by others, leading to an increased risk of accidents \cite{hri9, hri10}.

As risk-taking behaviour is mediated by the presence and encouragement of others, we set out to study whether similar effects can be observed in HRI. This study aims to investigate whether \emph{a robot has a similar impact on risk-taking behaviour as peer influence does in human-human interactions, and to understand the impact of physical and haptic interaction on risk-taking}. Previous research already confirmed that a robot's verbal encouragement can influence human risk-taking behaviour in participants \cite{hri2,kennedy2014children,vollmer2018children, salomons2018humans}.

In the studies reported here, we focus on the impact of affective touch and touch intensity on risk-taking behaviour, by measuring physiological arousal levels through skin conductivity (SC) and heart rate variability (HRV). Additionally, we investigate the effect of affective touch on trust
. Overall, our study aims to contribute to our understanding of the impact of affective touch in HRI on the user's physiological response and behaviour. 

This is to an extent related to trust. Trust is an increasingly important aspect of human-robot interactions (HRI) as interactions become more complex and have more consequential outcomes. Previous research has shown that social presence, including human tactile interaction, can enhance trust between humans and robots\cite{wang2019influence,law2021touching}.

\subsection{Social nature of tactile interaction}
The social nature of tactile interaction refers to the interpersonal and communicative aspects associated with physical touch between individuals. It encompasses the way touch is used to convey emotions, intentions, or establish social connections. Touch serves as our primary non-verbal means of expressing intimate emotions and is crucial for our physical and emotional well-being \cite{van2015social}. In today's digital era, much of our social interaction occurs through technology, but current communication tools like video calls lack the ability to convey touch sensations. This limitation means that digital communication cannot replicate the deep emotional connections formed through physical touch. To incorporate social touch into technology, systems need to not only generate touch sensations but also interpret \cite{ren2023low} and respond to touch. Social touch can have various effects on the receiver, including physiological, psychological, behavioural, and social impacts, and these effects often interact with one another. For example, a social presence and emotions can mutually reinforce each other during social touch interactions \cite{van2015social}. In the second study, we sought to determine whether the observed effects of the first study are primarily influenced by the social nature of tactile interaction or by the physical interaction itself.

The following section describes the human-robot interaction experimental design, including the research materials, experimental procedures, and data collection procedures. The data processing
methodology and performance metrics used in the study are introduced in Section~\ref{sec3:results_analysis}. To ascertain the social nature of touching a robot, we replicate the human-robot interaction study, but instead with a non-social object in Section~\ref{sec4:hoi_ex}. Results are presented together with a discussion in Section~\ref{sec:discussion}, along with reflections on the limitations of our research.

\section{Experimental design}
\label{sec2:Experimental_design}

The goal of the present study is to investigate the impact of tactile interaction on the physiological stress response, on risk-taking behaviour and on trust in the context of HRI. We present two studies. The first study examines the effects of different intensities of tactile interaction on stress responses and risk-taking behaviour. The second study determines whether the effects observed in the first study are caused by the \emph{social nature} of the tactile interaction or by the physical interaction alone. Our research answers the following research questions:

\begin{figure*}
\centering
\captionsetup{justification=centering}
\setlength{\abovecaptionskip}{0.2cm}
\begin{subfigure}{0.22\textwidth}
  \centering
  \includegraphics[width=\linewidth]{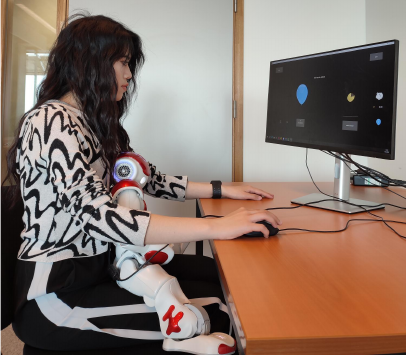}
  \caption{High-intensity\\ condition}
  \label{fig:high}
\end{subfigure}%
\hfill
\begin{subfigure}{0.22\textwidth}
  \centering
  \includegraphics[width=\linewidth]{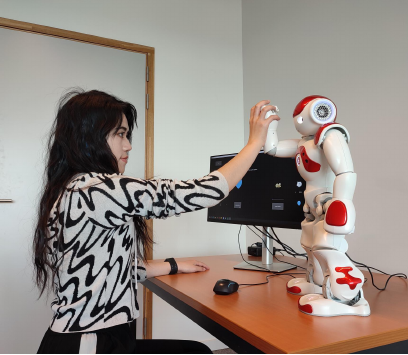}
  \caption{Low-intensity \\condition}
  \label{fig:low}
\end{subfigure}%
\hfill
\begin{subfigure}{0.22\textwidth}
  \centering
  \includegraphics[width=\linewidth]{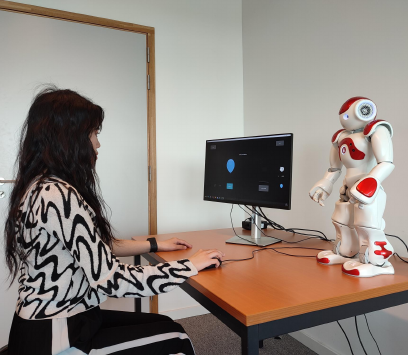}
  \caption{No-touch \\conditon}
  \label{fig:notouch}
\end{subfigure}%
\hfill
\begin{subfigure}{0.22\textwidth}
  \centering
  \includegraphics[width=\linewidth]{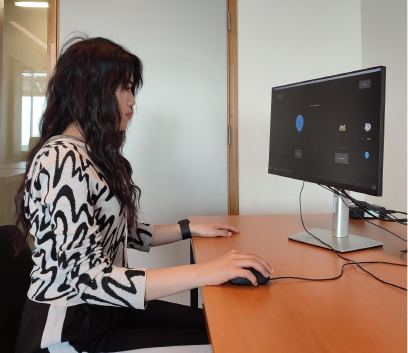}
  \caption{No robot \\ condition}
  \label{fig:norobot}
\end{subfigure}

\caption{Four conditions in the human-robot tactile interaction experiment (Study 1).}
\label{F16}
\end{figure*}

\begin{enumerate}
    \item Does tactile interaction with a social robot influence people's \emph{physiological responses}?
    \item Does tactile interaction with a social robot influence their \emph{risk-taking behaviour}?
    \item Does tactile interaction influence \emph{trust in the social robot}?
    \item Is it the \emph{social nature of the tactile interaction} or the \emph{physical interaction alone} that mediates attitudes and behaviour?
    \item Does tactile interaction influence risk-taking behaviour and stress levels differently across \emph{gender}?
    
\end{enumerate}

The results provide insight into the potential benefits and drawbacks of incorporating tactile interaction in HRI design and inform the design and implementation of robots in various settings such as healthcare, education, and industry. Furthermore, these experiments contribute to a deeper understanding of the role of tactile interaction in HRI and its potential impact on human behaviour. 


\subsection{Methodology and materials}\label{sec1}

The study uses a Balloon Analogue Risk Taking (BART) task, consistent with the methodology outlined in \cite{hri2}, to evaluate participants' risk-taking behaviour in a human-robot interaction and human-object interaction context. In the BART task participants play a computer game in which they inflate 30 balloons, one at a time. Participants can earn more by inflating the balloon, but risk losing their earnings when the balloon pops. At any time, they are given the option to collect their earnings by clicking the ``collect the money'' button, before proceeding to the next balloon. 

In the baseline condition, participants complete the BART task alone, as shown in Fig. \ref{F16}\subref{fig:norobot}. In the experimental condition, shown in Fig. \ref{F16}\subref{fig:high}-\subref{fig:notouch}, they complete the BART task in the presence of a robot. The robot encourages them to take more risk, by verbally nudging the participants to inflate the balloon a little more before moving on to the next balloon. 

In the first study, participants are in the presence of a small humanoid robot --- a United Robotics Group's Nao, a small 56cm humanoid robot--- and are invited to have tactile interaction with the intensity of the tactical interaction depending on the condition. In the second study, the robot is replaced by a non-social device (plastic box) to study whether the social character of the robot is a determining factor for the participant's response.

The study was run in accordance with the ethics guidelines of Ghent University and received approval from the Faculty of Engineering and Architecture based on a self-assessment of the research risks.

\subsubsection{Study 1: Human-robot interaction}\label{sec1}

In study 1 the participants are exposed to four experimental conditions: low-intensity robot touch interaction, high-intensity robot touch interaction, non-robot touch interaction, and no robot interaction, as depicted in Fig.~\ref{F16}.
The BART task was used in all conditions.


\begin{figure*}
\centering
\captionsetup{justification=centering}
\setlength{\abovecaptionskip}{0.2cm}
\begin{subfigure}{0.3\textwidth}
  \centering
  \includegraphics[width=\linewidth]{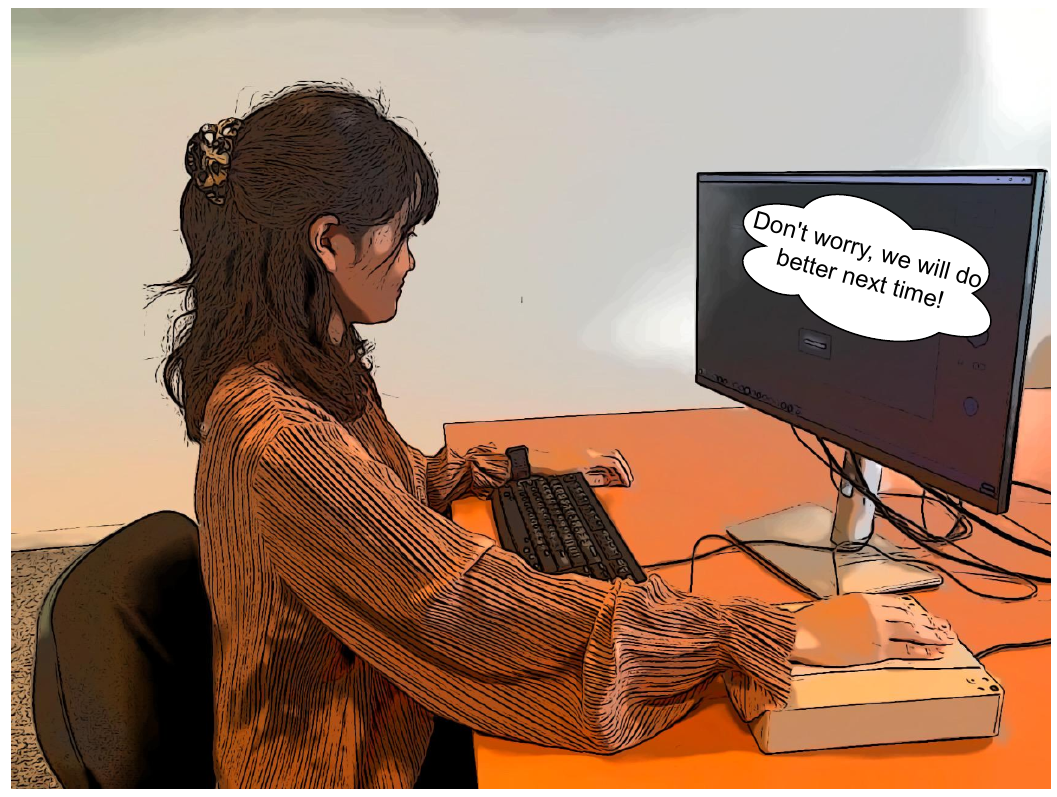}
  \caption{Affective tactile interaction with the box}
  \label{fig:affective}
\end{subfigure}%
\hfill
\begin{subfigure}{0.31\textwidth}
  \centering
  \includegraphics[width=\linewidth]{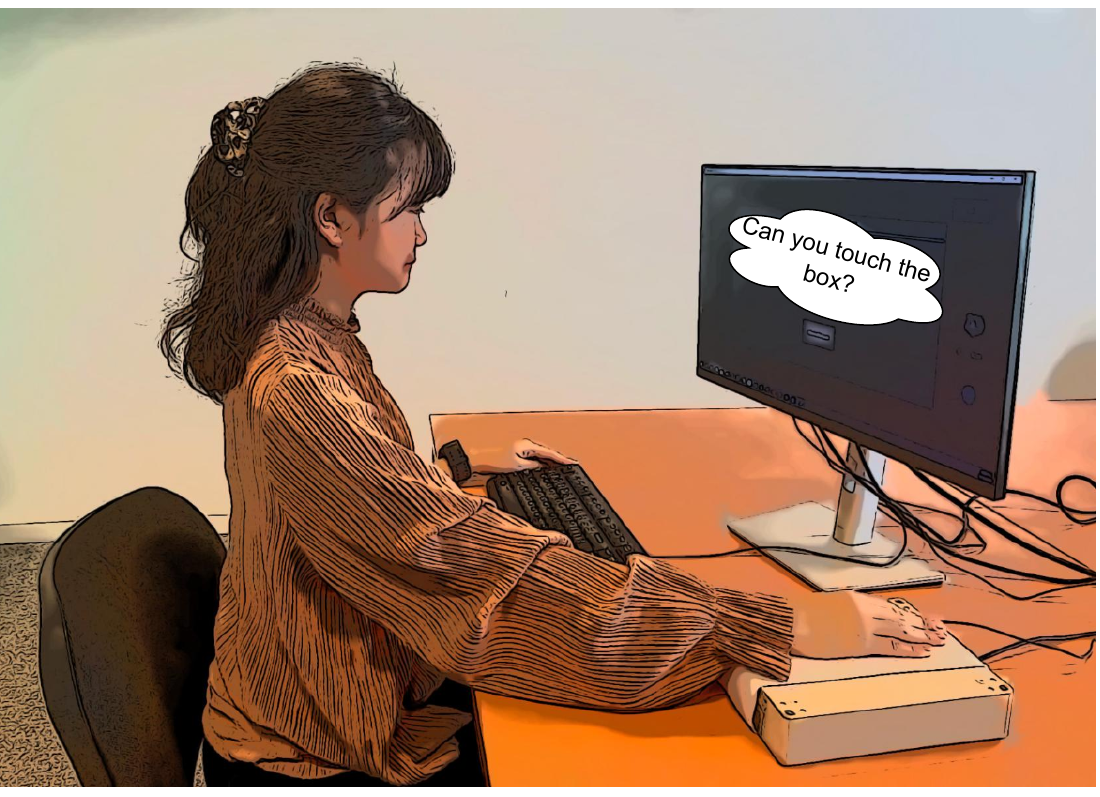}
  \caption{Neutral tactile interaction with the box}
  \label{fig:neutral}
\end{subfigure}%
\hfill
\begin{subfigure}{0.3\textwidth}
  \centering
  \includegraphics[width=\linewidth]{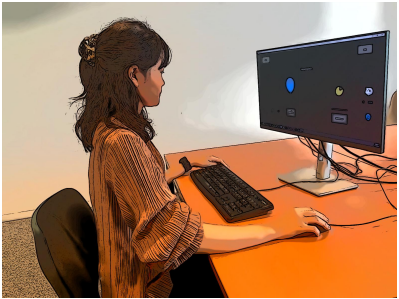}
  \caption{No tactile \\ interaction}
  \label{fig:noobject}
\end{subfigure}%

\caption{Three conditions in the human-object tactile interaction experiment (Study 2).}
\label{F17}
\end{figure*}

\begin{enumerate}
    \item In the \emph{low-intensity condition (LI)}  the robot invites the participant to have brief and gentle tactile interactions with the robot, timed to coincide with the participant wanting to move on to the next balloon  before they reached 50 pumps at which time the robot encourages them to further inflate the balloon or at the time balloon exploded. Participants are invited by the robot to shake hands, give a high-five or touch the robot's head, as a means to establish an affective tactile interaction. The encouragements to take more risk are taken from Hanoch \emph{et al.}\cite{hri2}.   

    \item In the \emph{high-intensity condition (HI)} the participants are instructed to take the robot on their lap to establish an intense and prolonged tactile interaction. During this, both the robot and participant face the screen. The robot's gestures and spoken utterances are the same as in the low-intensity condition. 
    
    \item In the \emph{no-touch condition (NT)} the participants are verbally encouraged by the robot, the robot uses co-speech gestures to accentuate its speech. The robot's encouragements are timed to coincide with the participant wanting to move on to the next balloon --at which time the robot encourages them to further inflate the balloon-- or at the time the balloon explodes. Participants do not physically interact with the robot.
    
    \item The \emph{no robot condition (NR)} serves as a baseline to compare the former three conditions to. In this condition, there is no robot present and no encouragement is given to take more risks. 


\end{enumerate}

\subsubsection{Study 2: Human-object interaction}\label{sec1}

In study 2, participants are invited to complete the BART task while interacting with a non-social object: a plastic box having the same texture as the surface of a Nao robot, as shown in the Fig.~\ref{F17}, to elicit similar tactile sensations. The box measured $23cm\times15cm\times5cm$ and is equipped with a capacitance pressure sensor to detect a touch event. We use two experimental conditions during which participants complete a BART task, as well as one control condition, as depicted in Fig.~\ref{F17}.

\begin{enumerate}

\item \emph{Affective tactile interaction (AT)} with the object. Participants are asked to tap the box by a high-valence message displayed on screen, e.g. ``Don’t worry! Could you touch the box? We will do better next time.'' These instructions are timed to coincide with the participant wanting to move on to the next balloon or with a balloon popping. 

\item \emph{Neutral tactile interaction (NAT)} with the object. Participants have low-intensity tactile interactions with the box, they are instructed to touch the box by showing message on the screen. This message is neutral in valance - ``Can you touch the box?''. The instructions are displayed when participants want to move on to the next balloon.

\item \emph{No tactile interaction condition (NT)}: In this condition, only affective sentences like -``you can do better next time'' are shown on the screen to encourage the participant to take more risks. This condition serves as a baseline. 

\end{enumerate}

\subsection{Participants}\label{sec1}

Participants who met the exclusion criteria (self-reported difficulties with reading or hearing, or reported acute or chronic heart conditions, as these conditions could potentially impact the ability to collect appropriate data) are excluded from the data collection. The data collection and study were conducted in accordance with the ethics regulations of the \emph{Universiteit Gent}. All participants provided informed consent and only took part in one of the two studies.

\begin{enumerate}

\item \emph{Study 1: Human-robot interaction experiment.} 38 participants (19 identified as cisgender males and 19 as cisgender females; mean age of 27.0 ± 2.2 years old) were recruited through a local social media campaign and were offered a 5 \euro\ voucher for an online store. Participants were randomly assigned to start with one of four experimental conditions ---the low-intensity touch condition (LI), the high-intensity touch condition (HI), the no-touch condition (NT), and the no-robot condition (NR)--- and saw the other conditions in balanced random order. The sample size calculation was performed using G*Power \cite{hri3}. The results of the calculation indicated that a sample size of 38 individuals with $\alpha = 0.05$ and an assumed effect size of 0.25 achieved an adequate statistical power of 96.2\% for the study.

\item \emph{Study 2: Human-object interaction experiment.} 36 participants (16 cisgender female and 20 cisgender male, age: 27.2 ± 2.5) took part. A power calculation indicated that the statistical power 90.6\%, assuming effect size $= 0.25$, $\alpha = 0.05$. All participants saw all three conditions --—NT, AT, and NAT conditions—-- in random balanced order.

\end{enumerate}

\subsection{Procedure}\label{sec1}

Upon arrival, participants signed an informed consent form and were instructed to wear the Empatica E4 sensor on their non-dominant hand for 15 minutes to establish a baseline measure of electrodermal activity (EDA) prior to the experiment. The E4 sensor is capable of collecting both inter-beat interval (IBI) and EDA data. Participants were instructed to sit comfortably and relax in a quiet room during this period. They were told that they would play a game where the goal is to make as much profit as possible.


\begin{enumerate}

\item \emph{Study 1: Human-robot interaction experiment.} The experiment session lasted on average one hour and ten minutes. After collecting baseline data, participants were asked to complete three questionnaires: the Multi-Dimensional Measure of Trust version 2 (MDMT v2) questionnaire, the Negative Attitude toward Robots Scale (NARS) questionnaire, and a self-reported risk-taking questionnaire. 
Upon completing the questionnaires, participants saw all four conditions in a random balanced order. After the first interaction with the robot, participants were asked to complete the MDMT v2 again (as this is the first and only condition in which they meet the robot without being influenced by prior encounters). and fill out two additional questionnaires: ``Participants' perceptions of the tactile interaction with Nao'' and ``Tactile interaction with Nao on the participants' emotional state.'' Measurements of IBI and EDA were sampled continuously during periods P1, P2, P3, and P4, the BART task's four conditions. Baseline IBI and EDA data are obtained at P0, before the task start. The time points T0, T1, T2, T3, and T4 refer to the end of the interaction in the baseline, first, second, third, and fourth periods, respectively. During P0 we ask participants to sit quietly for 15 minutes to obtain a physiological baseline measurement. After the first interaction, the participants complete the above-mentioned questionnaire, which takes 5 to 12 minutes to finish. There is one minute of breaks between the conditions. The first two minutes of physiological data in the P1, P2, P3, and P4 periods were not included in the analysis to prevent the previous interaction influencing the current one. The entire experiment lasted on average 1 hour and 10 minutes. Specifically, LI, HI, NR, and NT conditions lasted on average around 15 minutes, 10 minutes, 6 minutes, and 8 minutes, respectively.

\item \emph{Study 2: Human-object interaction experiment.} This study lasted an average of 45 minutes. Specifically, AT, NAT, and NT conditions lasted on average, 9 minutes, 9 minutes, and 6 minutes, respectively. Participants saw all three conditions in a random balanced order. IBI and EDA measurements were continuously recorded during the BART task in periods P1, P2, and P3, and the baseline of the IBI and EDA data was obtained at P0. There was a one-minute break between conditions. As before, the first two minutes of physiological data in the P1, P2, and P3 were not included in the analysis to prevent the impact of the previous interaction.

\end{enumerate}

Upon finishing, participants were debriefed and asked for feedback.

\subsection{Data analysis}

 The inter-beat interval (IBI) data were analyzed using the Kubios HRV standard software and further analysis using IBM SPSS V28. Parametric and nonparametric analyses of variance with corrected post hoc tests were used to evaluate the effect of the experimental conditions. EDA data were processed using the Ledalab toolbox within Matlab. The normality of the distributions is assessed using the Shapiro-Wilk test, with the significance level set at 0.05. Any identified outliers that were more than 5 standard deviations away from the mean were removed from the dataset. Additionally, to ensure the validity of the statistical inferences drawn from the data, the assumption of Sphericity was confirmed through the application of Mauchly's Test of Sphericity, which was conducted at a significance level of 0.05. The results of the Mauchly's W test and the corresponding p-values are reported in Table~\ref{tab4}. For multiple comparisons, we conducted post hoc analyses using Tukey’s HSD test for between-subjects factors. For within-subject factors, we applied the Bonferroni correction to control for family-wise error rates due to multiple comparisons. The Bonferroni correction was chosen for within-subjects factors to account for potential violations of sphericity and to maintain a conservative control over Type I error across the correlated tests inherent in repeated measures \cite{lee2018proper, maxwell1980pairwise}. 
  



\subsection{Equipment}\label{sec1}
\subsubsection{Nao Robot}
The Nao robot is a 58 cm tall and 5.6 kg programmable social humanoid, used in various application domains such as healthcare \cite{cespedes2021socially} and education \cite{vogt2019second}. While Nao is mostly used for non-tactile social interaction, it has been extended to allow affective tactile interaction to convey emotions to the robot \cite{andreasson2018affective}. However, in the studies reported here, we use a standard V5 Nao with no additional tactile sensory capabilities.

\subsubsection{MDMI V2}
The Multi-Dimensional Measure of Trust version 2 (MDMT v2) is a validated trust measurement tool designed to assess human-robot trust \cite{nomura2006measurement}. There are two broader factors of trust, consisting of five dimensions, which are Performance Trust (Reliable, Competent) and Moral Trust (Ethical, Transparent, Benevolent). The five dimensions consist of four items respectively, which are given below:

\begin{enumerate}
        \item Performance trust. 
        \begin{enumerate}
          \item Reliable Subscale: reliable, predictable, dependable, consistent
          \item Competent Subscale: competent, skilled, capable, meticulous
        \end{enumerate}
        \item Moral trust.
        \begin{enumerate}
          \item  Ethical Subscale: ethical, principled, moral, has integrity.
          \item Transparent Subscale: transparent, genuine, sincere, candid.
          \item Benevolent Subscale: benevolent, kind, considerate, has goodwill.
        \end{enumerate}
    \end{enumerate}
    
\subsubsection{NARS}

The Negative Attitude toward Robots Scale (NARS) is a measure of the negative attitudes and emotions of participants towards robots. It consists of three sub-scales ---negative attitudes toward situations of interaction with robots, negative attitudes toward the social influence of robots, and negative attitudes toward emotions in interaction with robots--- and has been widely used in HRI studies \cite{nomura2006altered}. 

\subsubsection{Empatica E4 sensor}

The Empatica E4 sensor is a wearable device that is equipped with various sensors for collecting physiological data, including a photoplethysmography (PPG) sensor, an EDA sensor, a 3-axis accelerometer, and an infrared thermopile. The E4 sensor is unique in that it is the only wearable sensor that combines EDA and PPG sensors, allowing for the simultaneous measurement of sympathetic nervous system (SNS) activity and skin conductance. It has been used in earlier studies, for example, to measure EDA and Emotional Response by E4 sensor \cite{chan2020effect}. In the studies reported here, the E4 sensor was worn on the non-dominant hand of the participants. They were instructed to refrain from moving that hand to avoid any spurious readings. This procedure was implemented to ensure that the physiological data collected were a direct result of the experimental manipulation and not due to extraneous factors.

\subsubsection{Balloon Analogue Risk Task}

The BART measures the propensity to take real-world risks. It is implemented as a game in which earnings are balanced against losses. In the task, participants inflate 30 different balloons by clicking a button. With each click, the balloon is inflated a little and 0.01 \euro\ is earned. Participants can choose to move on to the next balloon, at which time they collect the money earned for that balloon. If however the balloon is over-inflated and explodes, which happens after a pseudorandom number of pumps, the participant loses the money earned for that balloon. The larger the balloon, the greater the risk it will explode, but also the greater the potential reward. 
The BART results are complemented by a risk self-evaluation questionnaire.

\subsection{Measurements}\label{sec1}


\subsubsection{BART performance}

In the present study, various measures of performance on the BART were used to assess participants' risk-taking behaviour. The BART score, which is a commonly used measure in the literature, is calculated as the adjusted average number of pumps on unexploded balloons \cite{bornovalova2005differences,lejuez2002evaluation}. This score is a proxy for risk-taking behaviour, with higher scores indicating greater risk-taking propensity. In addition to the BART score, other standard measures of performance on the BART were also assessed, including (1) the total number of pumps by the participants, (2) the number of pumps at which a balloon explodes, (3) the number of exploded balloons, and (4) participants' profits for each balloon. These scores 
are the sum across the 30 trials. These measures have also been previously used in related literature \cite{hri2}.

\subsubsection{Gender difference in physiology and BART performance} 

There is a large body of literature that describes gender differences in risk-taking behaviour, with the consensus being that males are more likely to take risks than females, and that females are less likely to engage in activities which are competitive and have less predictable outcomes \cite{fletschner2010women,lazanyi2017analysis}. Studies have also shown that men tend to engage in more risky behaviour than women across various domains \cite{harris2006gender, charness2012strong} and that this gender difference is partially mediated by women paying more attention to potentially negative outcomes and experiencing lower satisfaction from risky tasks, and men are more aggressive, thrill-seeking and that they are willing to accept higher risks \cite{turner2003age}, and that young boys more readily engage in risk-taking behaviour \cite{ginsburg1982sex}. One potential explanation for these gender differences is that young men are more likely to be encouraged by their peers to engage in risky behaviour \cite{davies2000identifying}. Additionally, societal expectations, cultural and environmental factors, and cognitive and emotional differences could also play a role in these gender differences in risk-taking behaviour \cite{byrnes1999gender,harris2006gender,fessler2004angry}. However, it is worth noting that this is an active area of research and the literature is not conclusive, with further research being needed to fully understand the underlying causes of gender differences in risk-taking behaviour.

Based on this, we explore the gender differences in BART scores and HRV and the interaction between experimental conditions and gender. 

\subsubsection{Arousal level}

The robots' physical presence has been used to provide mental support in previous work \cite{shibata1999emergence}. In this context, arousal is often a key indicator and can be measured over longer intervals using physiological data, such as EDA and Heart Rate Variability (HRV) \cite{birenboim2019wearables}. For example, research has shown that remotely delivered touch can change the overall psychological affective experience as seen in the SCR \cite{chan2020effect}. In the present study, we measure arousal levels using EDA and HRV during the BART. Arousal is widely considered one of the two main aspects of emotional response, and while measuring arousal is not the same as measuring emotion, it is an important component of emotion \cite{caruelle2019use}. By measuring arousal, we aim to gain insights into how different intensities of tactile interaction affect emotional response and attention during the BART task.

\paragraph{Heart rate variability}

Heart rate variability (HRV) has been widely used as a measure to assess stress levels and emotional arousal, which measures the variations between heartbeats and is considered a proxy for stress \cite{thayer2012meta}. It can be analyzed using inter-beat interval (IBI) data recorded by the Empatica E4 sensor. However, stress levels can also be indirectly assessed through other measures, such as the Parasympathetic Nervous System (PNS), the sympathetic nervous system (SNS), and the Stress index. The Kubios HRV software can be used to perform HRV analysis by using IBI data, and it provides additional measures that are indicative of physiological elements connected to emotional regulation, such as the PNS index, the SNS index, and the stress (the square root of Baevsky's stress index) \cite{baevsky2008methodical}.

It is well established that HRV increases in response to PNS activity (vagal stimulation) and decreases in response to SNS activity \cite{hri7,hri8,hri6}. Stress also affects HRV, and current neurobiological research supports HRV as an objective indicator of psychological well-being and stress. When an individual is under stress or pressure, the high-frequency component of HRV is decreased \cite{Kim2018Stress}. Previous studies have found that HRV increases when people find continuous tactile stimulation pleasurable \cite{triscoli2017heart}. Lower PNS and higher SNS index values are associated with higher stress levels. In this study, participants' stress levels are measured during four different experimental conditions.

\paragraph{Electrodermal Activity}

Electrodermal Activity (EDA), also referred to as skin conductance, is a physiological response characterised by changes in the electrical conductance of the skin\cite{braithwaite2013guide}. These changes are triggered by external or internal stimuli and are thought to be indicative of emotional and physiological arousal \cite{nikula1991psychological}. EDA has been found to vary in relation to changes in attention, emotion, and motivational processes, making it a valuable tool for measuring an individual's emotional response. The use of EDA as an indicator of emotional response has been well-established in the literature. It is considered a valuable tool for understanding the physiological underpinnings of emotional responses, and for providing insights into the relationship between emotions, stress, and physiological responses \cite{figner2011using
,ventura2022new}. The use of EDA in the current study allows for the examination of the relationship between tactile interaction, emotional response, and physiological arousal in a controlled experimental setting.

In this study, we use Ledalab, following \cite{chan2020effect}, to examine the variations in significant skin responses triggered by tactile interaction compared to those without it. 
Ledalab is used to analyse the EDA data collected from the Empatica E4 sensor. Our analysis involves several steps. First, a low-pass filter was applied to minimize high-frequency noise. Subsequently, the Continuous Decomposition Analysis is employed to extract the signal properties of the underlying sudomotor nerve activity. By setting the threshold at 0.1, we identify significant peaks. To assess the correlation between touch occurrences and EDA peaks during low-intensity robot touch scenarios, we compare the timing of touch events with the timing of EDA peaks. The results revealed a 78\% coincidence, indicating that participants experience significant EDA peaks when they engage in tactile and emotion-laden interactions.

\subsubsection{Trust}

Trust is essential to interaction at different levels, from one-on-one interaction to large-scale societal interactions \cite{nataliya2015trust}. 
During human-robot interaction, the outcome of an interaction is often influenced by trust \cite{billings2012human}. Earlier studies point out that human contact has a positive effect on the formation of trust. The opposite behaviour, deception and malfunctioning, is equally important. When robots deceive people or make mistakes, it reduces the trust in the robot \cite{walter2013trust}. We use the MDMT v2 questionnaire to measure trust in the robot.

\subsubsection{Perception of the interaction of Nao}

We administer a questionnaire to assess participants' perceptions of Nao and their attitudes towards the robot after the first interaction condition. 
The questionnaire consisted of 8 statements on Nao's perceived feelings and 6 statements on the participant's feelings, inspired by the previous research \cite{geva2020touching}, using a 5-point Likert-scale 
(see the full questionnaire in Supplementary Materials). The questionnaire is completed after completing the first interaction condition.

\subsubsection{Relationship between trust, risk-taking behaviour and arousal} 

The relationship between trust and risk is not clear yet, even though they have been linked in the literature \cite{siegrist2021trust}. 
Trust is divided into perceived trust (also known as subjective trust) and behavioural trust. Perceived trust is inversely related to perceived risk, and behavioural trust is related to risk-taking, The relationship between subjective trust and behavioural trust parallels the relationship between perceived risk and the risk-taking behaviours \cite{das2004risk}. 
Our data also allow us to study whether risk-taking behaviour is related to the extent to which humans trust robots -- an intriguing and as yet unexplored issue. 

\section{Results and analysis}
\label{sec3:results_analysis}

\subsection{Perception of the interaction of Nao} 
 
\noindent \textbf{Participants' subjective experience of tactile interaction} 

\begin{figure}[H]
\centering
\includegraphics[width=\columnwidth{}]{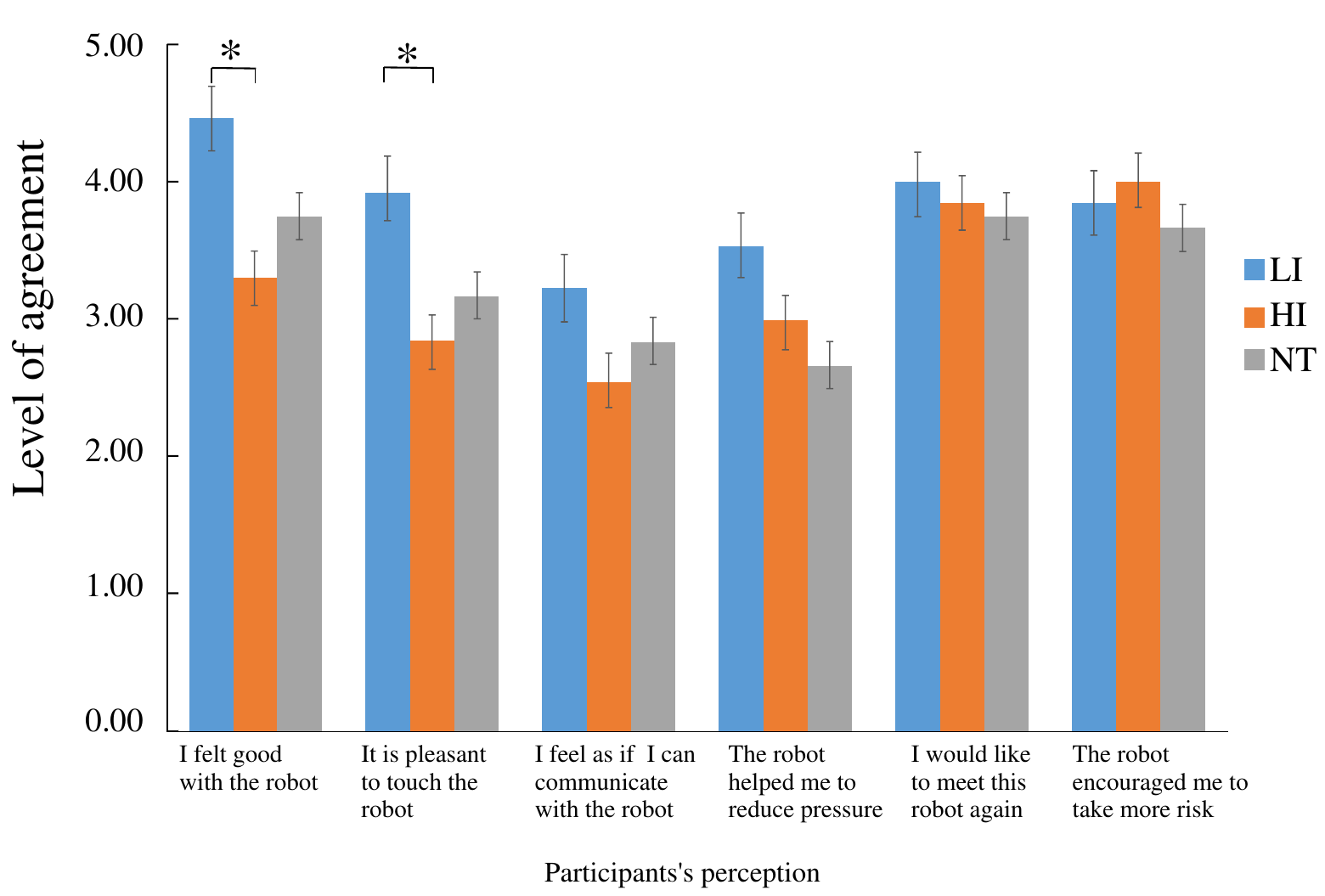}

\caption{\mybluehl{Participants’ attitudes after their first interaction with the robot, measured across 3 conditions: low-intensity tactile interaction (LI), high-intensity tactile interaction (HI) and no touch (NT). Bars show the average scores per statement, whiskers indicate Standard Error.} 
} \label{F3}
\end{figure}

Fig.~\ref{F3} shows the subjective perception of participants for different conditions. 
Note how Fig.~\ref{F3} 
shows that LI ranked highest among the three experimental conditions, except for `the robot encouraged me to take more risk'. The results of the one-way ANOVA revealed significant effects of the experimental conditions on participants' perceptions, particularly note on two items: `I felt good with the robot' ($F(2,35)=4.25, p = 0.022, \eta^2 = 0.195$), and `It is pleasant to touch the robot' ($F(2,35) = 4.05, p = 0.026, \eta^2 = 0.188$). Further post hoc analysis indicates that for the item `I felt good with the robot', participants in the LI condition reported that they felt significantly better with the robot than they did in the HI condition. ($p = 0.018$). Similarly, for the item `It is pleasant to touch the robot', participants reported significantly greater tactile enjoyment in the LI condition than in the HI condition ($p = 0.023$).

\noindent \textbf{Participants' perception of Nao's emotional state}

\begin{figure}[H]
\setlength{\abovecaptionskip}{0.2cm}
\setlength{\belowcaptionskip}{-0.4cm}
\centering
\includegraphics[width=\columnwidth]{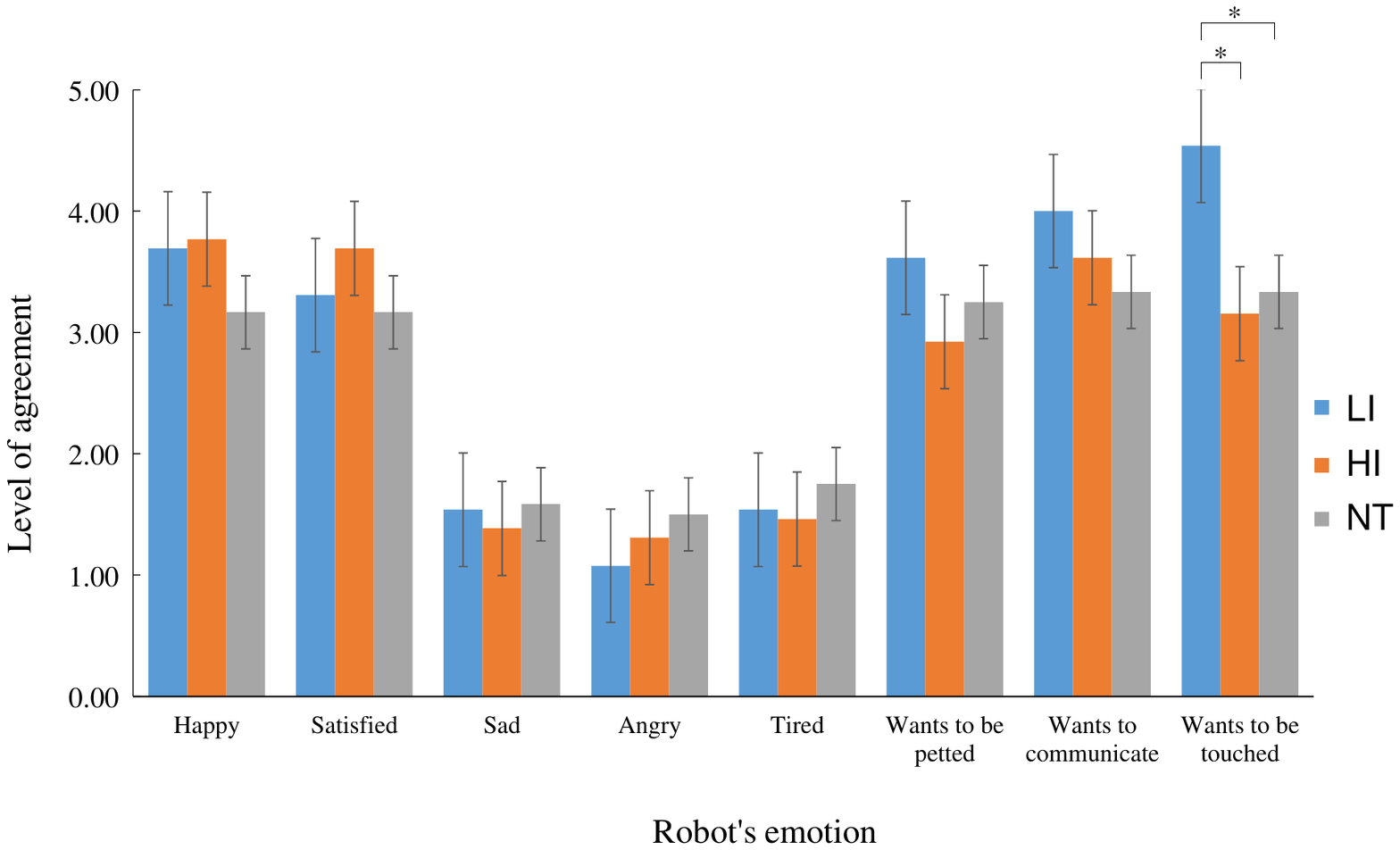}
\caption{Participants' perception of Nao’s emotional state. Bars show the average scores for each emotional state, whiskers indicate Standard Error. Participants perceived the robot to be experiencing positive emotions rather than negative emotions.} \label{F2}
\end{figure}

The subjective perception of Nao's emotional state is queried after the first interaction with the robot (at T1). As shown in the Fig.~\ref{F2}, 
high ratings are given to positive items: happy (average over all conditions 3.6±1.1), satisfied (3.4±0.9), wants to be petted (3.3±1.5), wants to be touched (3.7±1.5), and wants to communicate (3.7±1.2). Low ratings are given to Nao feeling tired (1.6±1.0), sad (1.5±0.6) and angry (1.3±0.7). Moreover, the result of one-way ANOVA shows a significant difference in participants' responses to the item `wants to be touched' across the experimental conditions ($F(2, 35) = 4.25, p = 0.022, \eta^2 = 0.227$). Post hoc tests show participants under the LI condition yield significantly higher scores on the `wants to be touched' item than the HI condition ($p = 0.018$). Furthermore, a significantly higher rating for the item `wants to be touched' is also observed in the LI condition than in the NT condition ($p = 0.033$).

\subsection{BART performance}

BART scores report the performance on the BART task, the number of pumps, the number of exploded balloons, and profits. Because of the small sample size ($N < 40$),  normality is checked using the Shapiro-Wilk test. As $p>0.05$ all measures can be considered as normal. 
The Shapiro-Wilk test and Mauchly's test results are reported in Table~\ref{tab3} and Table~\ref{tab4}, respectively.

\begin{table*}[t]\footnotesize 
\setlength{\abovecaptionskip}{0.0cm}   
	\setlength{\belowcaptionskip}{-0cm}  
	\renewcommand\tabcolsep{2.0pt} 
	\centering
	\caption{Test of Normality for BART scores and heart rate variability.}
	\begin{tabular}
	{
	p{2.5cm} 
 p{2cm}<{\centering} 
	 p{2.3cm}<{\centering} 
  p{2.2cm}<{\centering}
	p{2.2cm}<{\centering}
	} 
\hline
    
     {\textbf{Shapiro-Wilk}} & 
     {\textbf{HI}}  & 
     {\textbf{LI}} & 
     {\textbf{NT}} & 
     {\textbf{NR}}  \\
     
     \hline

   {BART Scores}&
  {$0.98, p>0.05$} &
 {$0.95, p>0.05$} &
 {$0.98, p>0.05$} &
 {$0.97, p>0.05$} \\


   {Explosions}&
  {$0.10,p>0.05$} &
 {$0.39,p>0.05$} &
 {$0.21,p>0.05$} &
 {$0.30,p>0.05$}\\
     
     {Pumps}&
  {$0.91,p>0.05$} &
 {$0.47,p>0.05$} &
 {$0.61,p>0.05$} &
 {$0.27,p>0.05$}\\
     
     {Stress index}&
  {$0.89, p>0.05$} &
 {$0.95, p>0.05$} &
 {$0.90, p>0.05$} &
 {$0.89, p>0.05$}\\
     \hline
     
	\end{tabular}
	\label{tab3}
\end{table*}

\subsubsection{BART scores}
A one-way repeated measures ANOVA results indicate significant differences in BART scores between the four interaction conditions ($F(3, 111) = 6.60, p < 0.001, \eta^2 = 0.151$). Post hoc tests revealed participants under HI ($p=0.042$) and LI ($p=0.037$) conditions score significantly higher than NT conditions, similarity, participants under HI ($p=0.024$) and LI ($p=0.005$) conditions significantly perform better than under NR conditions in scores. See Fig.~\ref{F10} and Table~\ref{tab5}, which show that both high-intensity and low-intensity tactile interaction with the robot can encourage people to take more risks.

\begin{table*}[h]\footnotesize 
\setlength{\abovecaptionskip}{0.0cm}   
\setlength{\belowcaptionskip}{-0cm}  
	\renewcommand\tabcolsep{1pt}  
	\centering
	\caption{Mauchly's Test of Sphericity for BART performance.}
	\begin{tabular}
	{p{3cm} 
 p{2cm}<{\centering} 
	 p{2.3cm}<{\centering} 
  p{2.2cm}<{\centering}
	p{2.2cm}<{\centering}} 
\hline 
     {\textbf{BART performance}} & 
     {\textbf{BART Scores}}  & 
     {\textbf{Explosions}} & 
     {\textbf{Pumps}} & 
     {\textbf{Profits}}  \\

    \specialrule{0em}{0.05em}{0.pt}
    \hline
		\specialrule{0em}{0.05em}{0.pt}
    
    
 \specialrule{0pt}{0pt}{0pt}
   {Mauchly's W}   &   {0.78, $p > 0.05$} &
  {0.85, $p > 0.05$} &
  {0.83, $p > 0.05$} &
  {0.97, $p > 0.05$}\\ 

     \hline
	\end{tabular}
	\label{tab4}
\end{table*}

\begin{figure}[H]
\setlength{\abovecaptionskip}{0.2cm}
\setlength{\belowcaptionskip}{-0.4cm}
\centering
\includegraphics[width=\columnwidth]{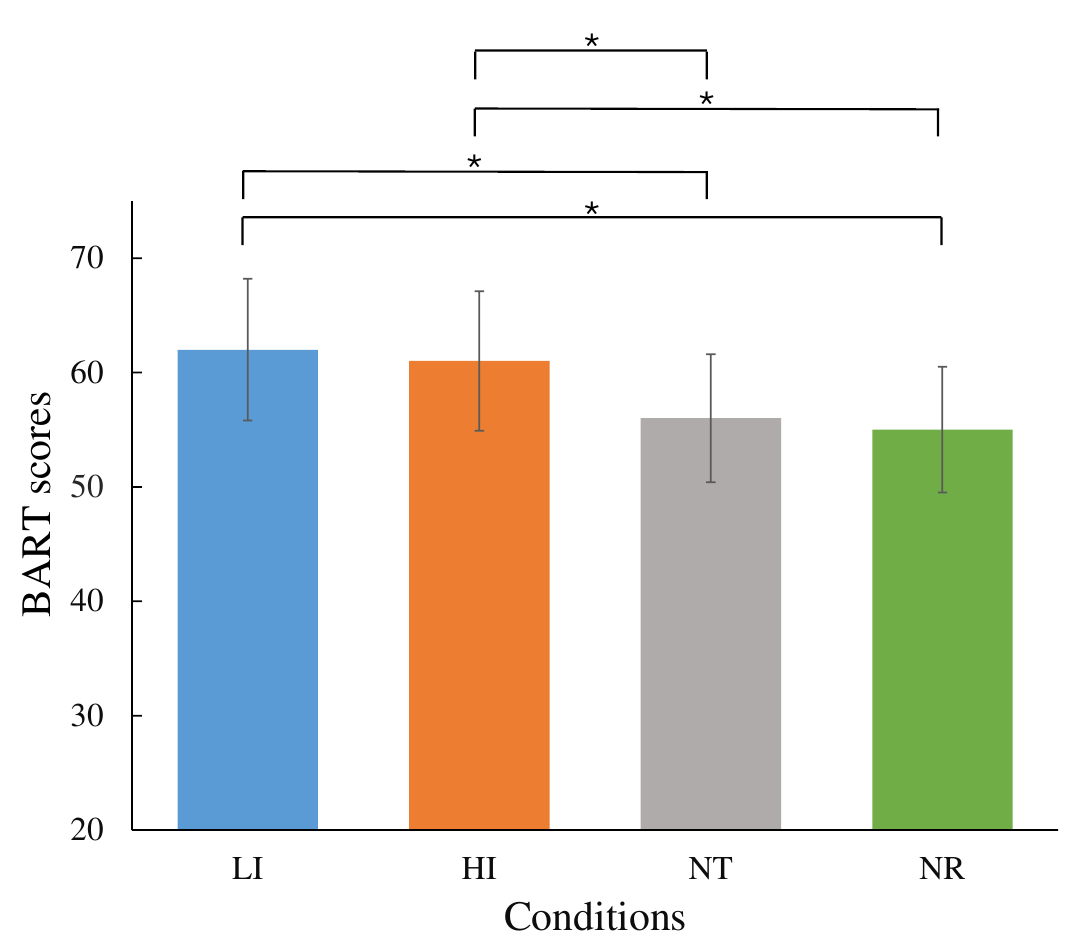}
\caption{\mybluehl{BART scores per condition. Bars show the mean scores for each condition (LI, HI, NT and NR conditions). Whiskers indicate Standard Error. (* $p < 0.05$)}. 
} \label{F10}
\end{figure}

\subsubsection{Number of Explosions}

\begin{figure}
\setlength{\abovecaptionskip}{0.2cm}
\setlength{\belowcaptionskip}{-0.4cm}
\centering
\includegraphics[width=\columnwidth]{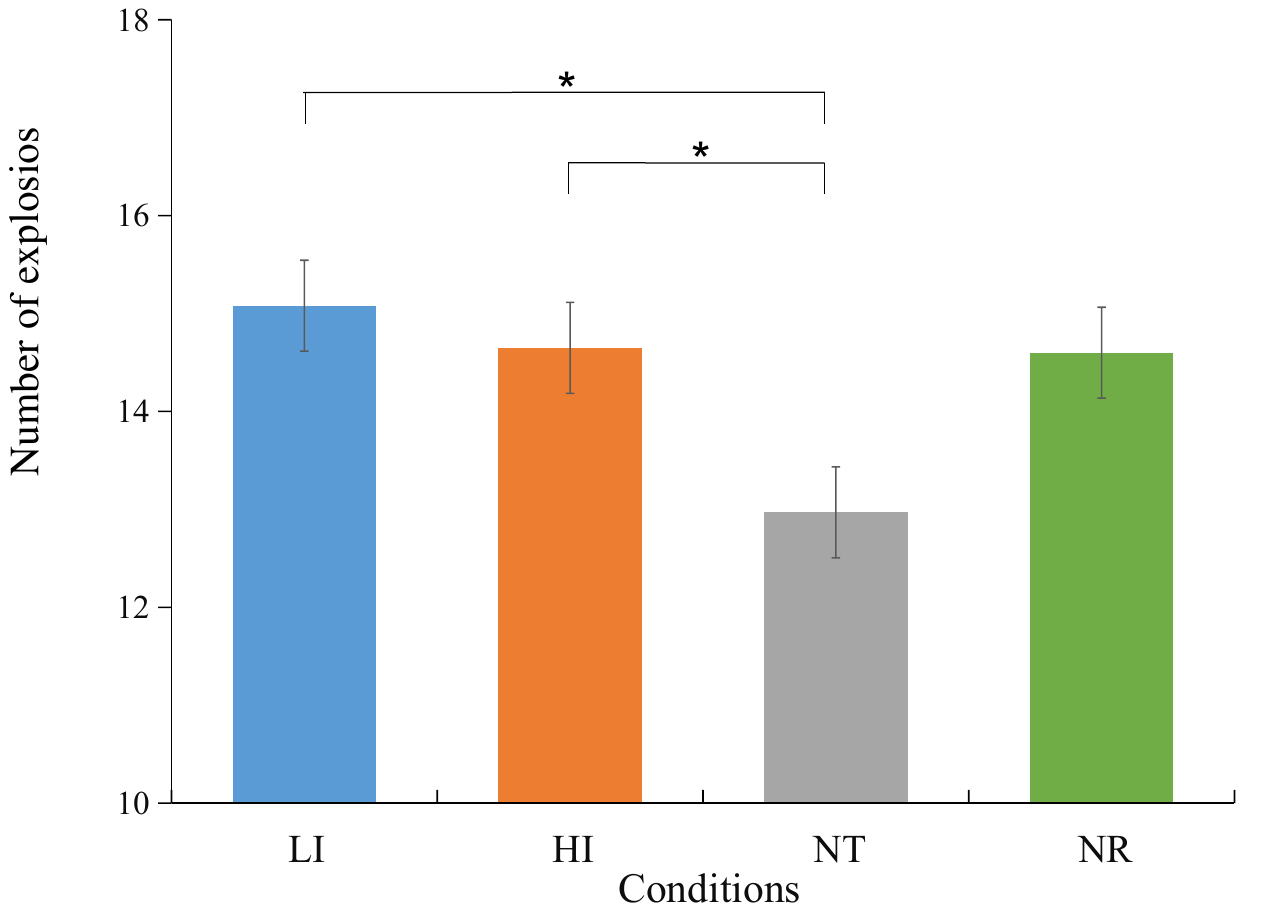}
\caption{Number of explosions per condition. Bars show the average number of explosions for each condition (LI, HI, NT and NR conditions), whiskers indicate Standard Error. (* $p < 0.05$). There is significant difference between LI, HI and NT conditions, respectively.} \label{F11}
\end{figure}

A significant main effect of the four experimental phases (HI/LI/NT/NR) is found for the number of explosions, as determined by a one-way repeated measures ANOVA ($F(3, 111) = 4.56, p = 0.005, \eta^2 = 0.112$). Post hoc tests indicate that the number of explosions is significantly higher in the Low-Intensity (p=0.012) and High-Intensity (p=0.045) conditions compared to the NT condition. However, there is no significant difference between NR and NT conditions, as seen in Fig.~\ref{F11} and Table~\ref{tab5}.

\subsubsection{Number of pumps}

\begin{figure}
\setlength{\abovecaptionskip}{0.2cm}
\setlength{\belowcaptionskip}{-0.4cm}
\centering
\includegraphics[width=1.1\columnwidth]{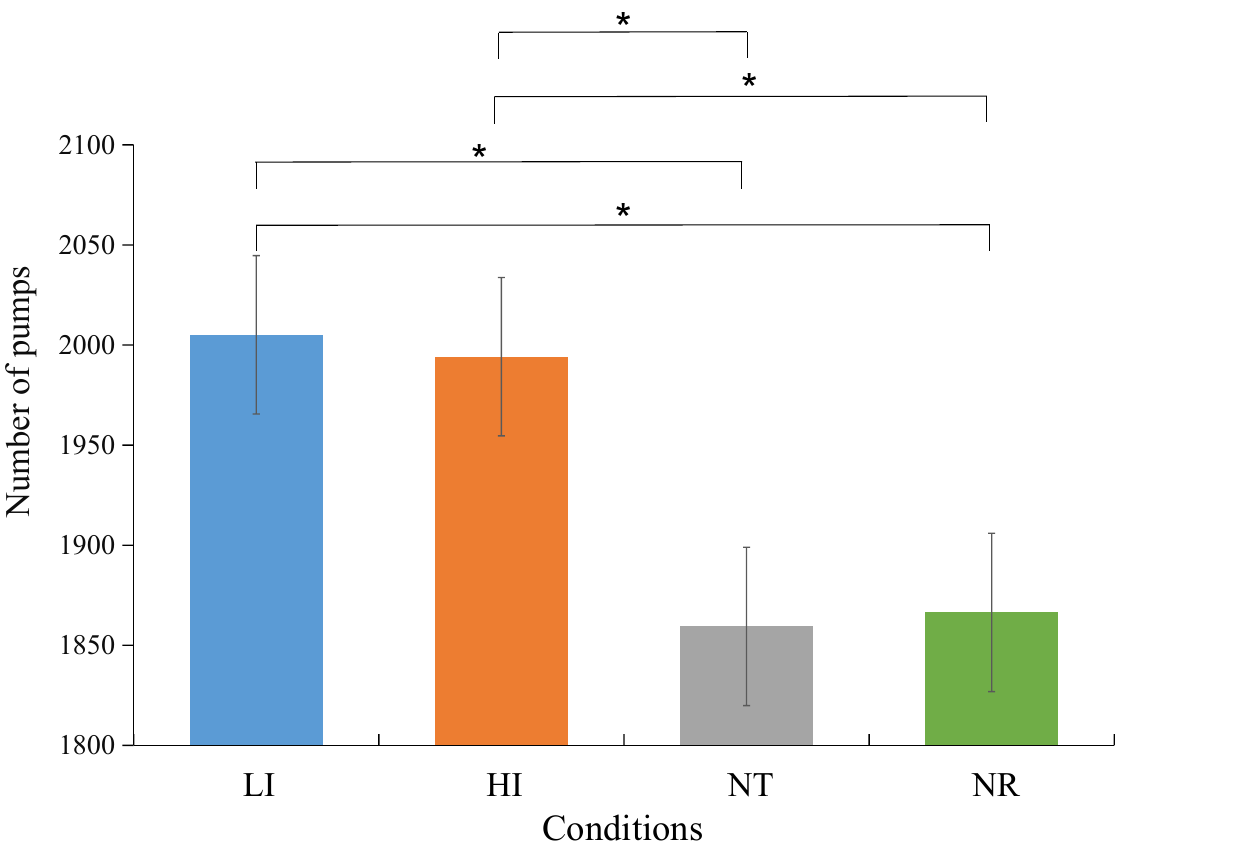}
\caption{Number of pumps per condition. Bars show the average number of pumps for each condition (LI, HI, NT and NR conditions), whiskers indicate Standard Error. (* $p < 0.05$). There is a significant difference between LI and NT conditions.} \label{F12}
\end{figure}

A one-way repeated measures ANOVA indicates a significant effect of conditions (HI/LI/NT/NR) for the number of total pumps ($F(3, 111) = 6.19, p < 0.001, \eta^2 = 0.147$). Post hoc tests demonstrate that the total number of pumps is significantly higher in the LI ($p = 0.014$) condition and HI ($p = 0.042$) condition than in the NT condition. In addition, participants in LI ($p = 0.047$) and HI ($p = 0.02$) conditions pump significantly more than in the NR condition, which can be seen in Fig.~\ref{F12}.

\subsubsection{Profits}

The study examines the profits generated by participants under four different experimental conditions (LI/HI/NT/NR). The profits are calculated to be $9.31\pm1.53$ \euro\ , $9.36\pm1.51$ \euro\ , $9.53\pm2.17$ \euro\  and $8.48\pm1.85$ \euro\  for the LI, HI, NT, and NR conditions, respectively ($p > 0.05$), as shown in Table \ref{tab5} and \ref{F13}. A one-way repeated measures ANOVA indicate that there is no significant difference in profits among the four conditions. However, profits are influenced by pumps and the number of exploded balloons. We can not deduce solely that the people who risked more with higher profit.  Because the highly risky people are prone to higher pumps and the higher number of exploded balloons, which means that profits might not be positively related to the people's risk-taking behaviour.

\begin{table*}[h!]\footnotesize 
\setlength{\abovecaptionskip}{0.0cm}   

	\setlength{\belowcaptionskip}{-0cm}  
	\renewcommand\tabcolsep{2.0pt} 
	\centering
	\caption{Means (and Standard Deviations) of BART performance by condition.}
	\begin{tabular}
	{p{2cm} 
 p{1.2cm}<{\centering} 
	 p{1.cm}<{\centering} 
  p{1.2cm}<{\centering}
	p{1.cm}<{\centering} 
 p{1.2cm}<{\centering}
	p{1cm}<{\centering} 
	p{1.2cm}<{\centering} 
 p{1cm}<{\centering}} 
\hline
 
        \multirow{2}{*}{\textbf{Variable}} &
        \multicolumn{2}{c}{\textbf{Low-intensity}} &
        \multicolumn{2}{c}{\textbf{High-intensity}} &
        \multicolumn{2}{c}{\textbf{No-touch}} &
        \multicolumn{2}{c}{\multirow{2}{*}{\textbf{No-robot}}} \\ 

        & \multicolumn{2}{c}{\textbf{touch}} & \multicolumn{2}{c}{\textbf{touch}} & \multicolumn{2}{c}{\textbf{touch}} & \\

        \specialrule{0pt}{0pt}{0pt}
        \hline
          \specialrule{0pt}{0pt}{0pt}
	\textbf{Metric}	& \textbf{M} & \textbf{SD}
  & \textbf{M} & \textbf{SD}  
  & \textbf{M} & \textbf{SD} 
  & \textbf{M} & \textbf{SD} \\
 
    \specialrule{0pt}{0pt}{0pt}
        \hline
          \specialrule{0pt}{0pt}{0pt}
     {BART Scores}  & 62 & 11 &
     61 & 12 &
     56 & 13 &
     55 & 14 \\

     {Explosions}  & 15.08  & 0.57  &
    14.65  & 0.46  &
     12.97 & 0.45 &
     14.60  & 0.54 \\

     {Pumps}  & 2005.1 & 205.8 &
    1994.2 & 167.4 &
    1859.4 & 246.6 &
    1866.4  & 189.6 \\

     {Profits}  & 9.31 & 1.53 &
     9.36 & 1.51 &
     9.53 & 2.17 &
     8.48 & 1.85 \\
     \hline
	\end{tabular}
	\label{tab5}
\end{table*}

\begin{figure}[h!]
\setlength{\abovecaptionskip}{0.2cm}
\setlength{\belowcaptionskip}{-0.4cm}
\centering
\includegraphics[width=\columnwidth]{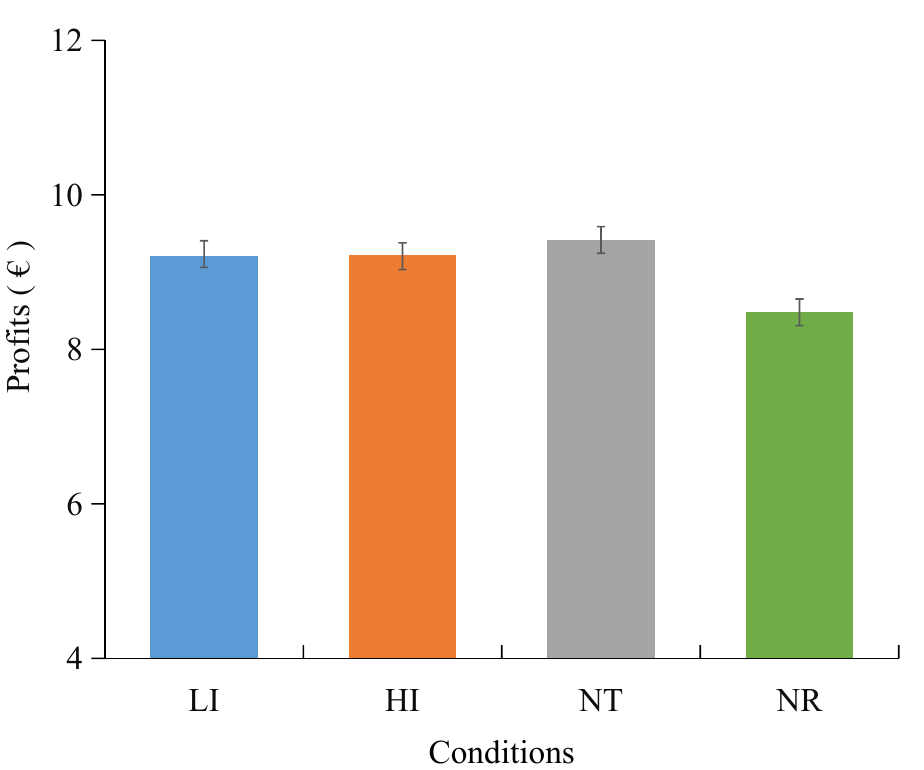}
\caption{Profits per condition. Bars show the average profits for each condition  (LI, HI, NT and NR conditions), whiskers indicate Standard Error. (* $p < 0.05$). No significant difference was found among all the conditions.} \label{F13}
\end{figure}

\begin{figure*}[h!]
\begin{minipage}[t]{0.5\textwidth}
\centering
\includegraphics[width=\textwidth]{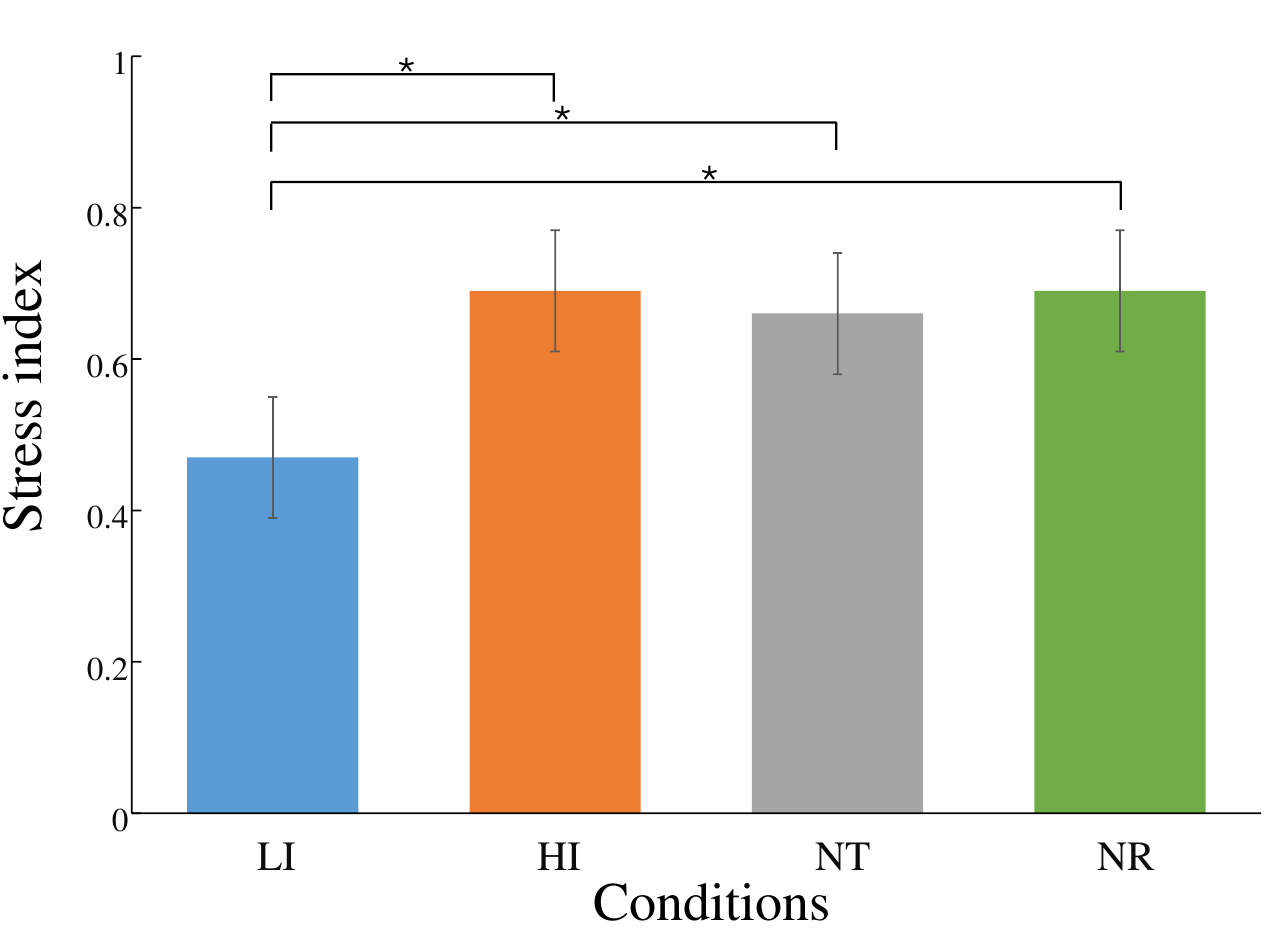}
\caption{\mybluehl{Stress index per condition. Bars show the average stress index for each condition  (LI, HI, NT and NR conditions). Whiskers indicate Standard Error. (* $p < 0.05$)}} \label{F9}
\end{minipage}
      \hspace{5mm} 
\begin{minipage}[t]{0.5\textwidth}
\centering
\includegraphics[width=0.9\columnwidth]{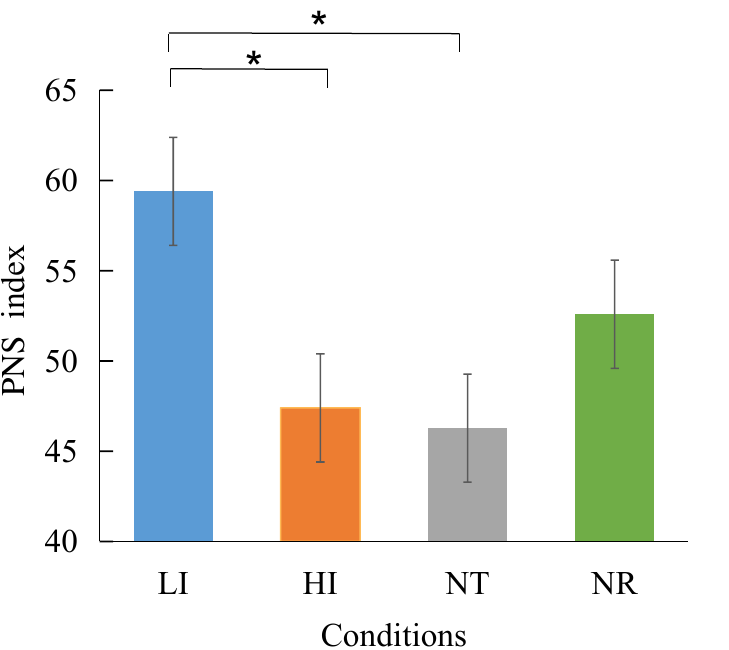}
\caption{PNS index per condition. Bars show the average PNS index for each condition  (LI, HI, NT and NR conditions), whiskers indicate Standard Error. (* $p < 0.05$). 
} \label{F14}
\end{minipage}
\end{figure*}


\subsection{Heart rate variability}

The results of the study, as presented in Fig.\ref{F9}, Fig.\ref{F14}, and Table~\ref{tab6}, indicate that the lowest stress levels are reported in the LI condition, as reflected by the highest PNS index value. The Stress index values show that the participants experienced higher levels of stress in the HI and NT conditions. The one-way repeated measures ANOVA revealed a significant difference in the PNS index among the four conditions ($F(3,111) = 5.88, p = 0.001, \eta^2 = 0.137$), and post hoc tests revealed significant differences between the LI and HI condition and the LI and NT condition, the LI condition exhibited significantly higher PNS index values than both the NT ($p = 0.001$) and HI ($p = 0.006$) conditions. In terms of the Stress index, a significant effect of the condition is observed ($F(3,111) = 7.07, p < 0.001, \eta^2 = 0.160$). Post-hoc tests highlight that the LI condition has a notably lower Stress index than the HI ($p < 0.001$), NT ($p = 0.006$), and NR ($p = 0.005$) conditions. These findings suggest that participants experienced the least pressure in the LI condition, while they reported higher stress levels in HI and NT conditions.

\begin{table*}
\footnotesize 
\setlength{\abovecaptionskip}{0.0cm}   
	\setlength{\belowcaptionskip}{-0cm}  
	\renewcommand\tabcolsep{5.0pt} 
	\centering
	\caption{Means (and Standard Deviations) of stress index, PNS index and SNS index by condition.}
	\begin{tabular}
	{
	p{2.0cm}
 p{1.5cm}<{\centering} 
	 p{1.2cm}<{\centering} |
  p{1cm}<{\centering}
	p{1.2cm}<{\centering} |
 p{1.2cm}<{\centering}
	p{1.3cm}<{\centering} |
	p{1.2cm}<{\centering} 
 p{1.3cm}<{\centering} 
	} 
	    \specialrule{0pt}{0pt}{0pt}

\hline

    \specialrule{0pt}{0pt}{0pt}

        {\textbf{Variable}} &
        \multicolumn{2}{c}{\textbf{High-intensity touch}} &
        \multicolumn{2}{c}{\textbf{Low-intensity touch}} &
        \multicolumn{2}{c}{\textbf{No-touch}} &
        \multicolumn{2}{c}{\textbf{No-robot}} \\

\hline

     {Stress index}  & 
      \multicolumn{2}{c}{$0.69\pm0.29$}  &
      \multicolumn{2}{c}{$0.47\pm0.12$}  &
      \multicolumn{2}{c}{$0.66\pm0.33$}  &
      \multicolumn{2}{c}{$0.69\pm0.35$} \\

     {PNS index}  & 
      \multicolumn{2}{c}{$47.40\pm3.24$}  &
      \multicolumn{2}{c}{$59.40\pm3.37$}  &
      \multicolumn{2}{c}{$46.28\pm3.49$}  &
      \multicolumn{2}{c}{$52.59\pm4.71$} \\

     {SNS index}  & 
      \multicolumn{2}{c}{$-2.83\pm0.20$}  &
      \multicolumn{2}{c}{$-3.09\pm0.11$}  &
      \multicolumn{2}{c}{$-2.79\pm0.18$}  &
      \multicolumn{2}{c}{$-3.01\pm0.19$} \\
     \hline

	\end{tabular}
	\label{tab6}
\end{table*}

\begin{table*}[h!]\footnotesize 
\setlength{\abovecaptionskip}{0.0cm}   

	\setlength{\belowcaptionskip}{-0cm}  
	\renewcommand\tabcolsep{2.0pt} 
	\centering
	\caption{Means (and Standard Deviations) of BART scores and stress index by gender.}
	\begin{tabular}
	{p{2cm} 
 p{1.5cm}<{\centering} 
	 p{1.5cm}<{\centering} 
  p{1.5cm}<{\centering}
	p{1.5cm}<{\centering} 
 p{1.5cm}<{\centering}
	p{1.5cm}<{\centering} 
	p{1.5cm}<{\centering} 
 p{1.5cm}<{\centering}} 
\hline
 
        \multirow{2}{*}{\textbf{Variable}} &
        \multicolumn{2}{c}{\textbf{Low-intensity}} &
        \multicolumn{2}{c}{\textbf{High-intensity}} &
        \multicolumn{2}{c}{\textbf{No-touch}} &
        \multicolumn{2}{c}{\multirow{2}{*}{\textbf{No-robot}}} \\ 

        & \multicolumn{2}{c}{\textbf{touch}} & \multicolumn{2}{c}{\textbf{touch}} & \multicolumn{2}{c}{\textbf{touch}} & \\

        \specialrule{0pt}{0pt}{0pt}
        \hline
          \specialrule{0pt}{0pt}{0pt}
	\textbf{Gender}	& \textbf{Female} & \textbf{Male}
 & \textbf{Female} & \textbf{Male}  
 & \textbf{Female} & \textbf{Male}
 & \textbf{Female} & \textbf{Male} \\
 
    \specialrule{0pt}{0pt}{0pt}
        \hline
          \specialrule{0pt}{0pt}{0pt}
     {BART Scores}  &  $56 \pm 11$ &  $68 \pm 8$ &
     $56 \pm 13$ &  $66 \pm 8$ &
      $50 \pm 14$ &  $61 \pm 11$ &
      $47 \pm 10$ &  $63 \pm 12$ \\

     {Stress Index}  &  $0.50 \pm 0.12$  &  $0.45 \pm 0.12$  &
    $0.81 \pm 0.33$  &  $0.57 \pm 0.21$ &
     $0.73 \pm 0.34$ &  $0.59 \pm 0.31$ &
      $0.81 \pm 0.41$  &  $0.58 \pm 0.22$ \\
     \hline

	\end{tabular}
	\label{gender}
\end{table*}

\subsection{Gender differences in risk-taking performance and stress level during BART task}



\begin{figure}
\setlength{\abovecaptionskip}{0.2cm}
\setlength{\belowcaptionskip}{-0.4cm}
\centering
\includegraphics[width=\columnwidth]{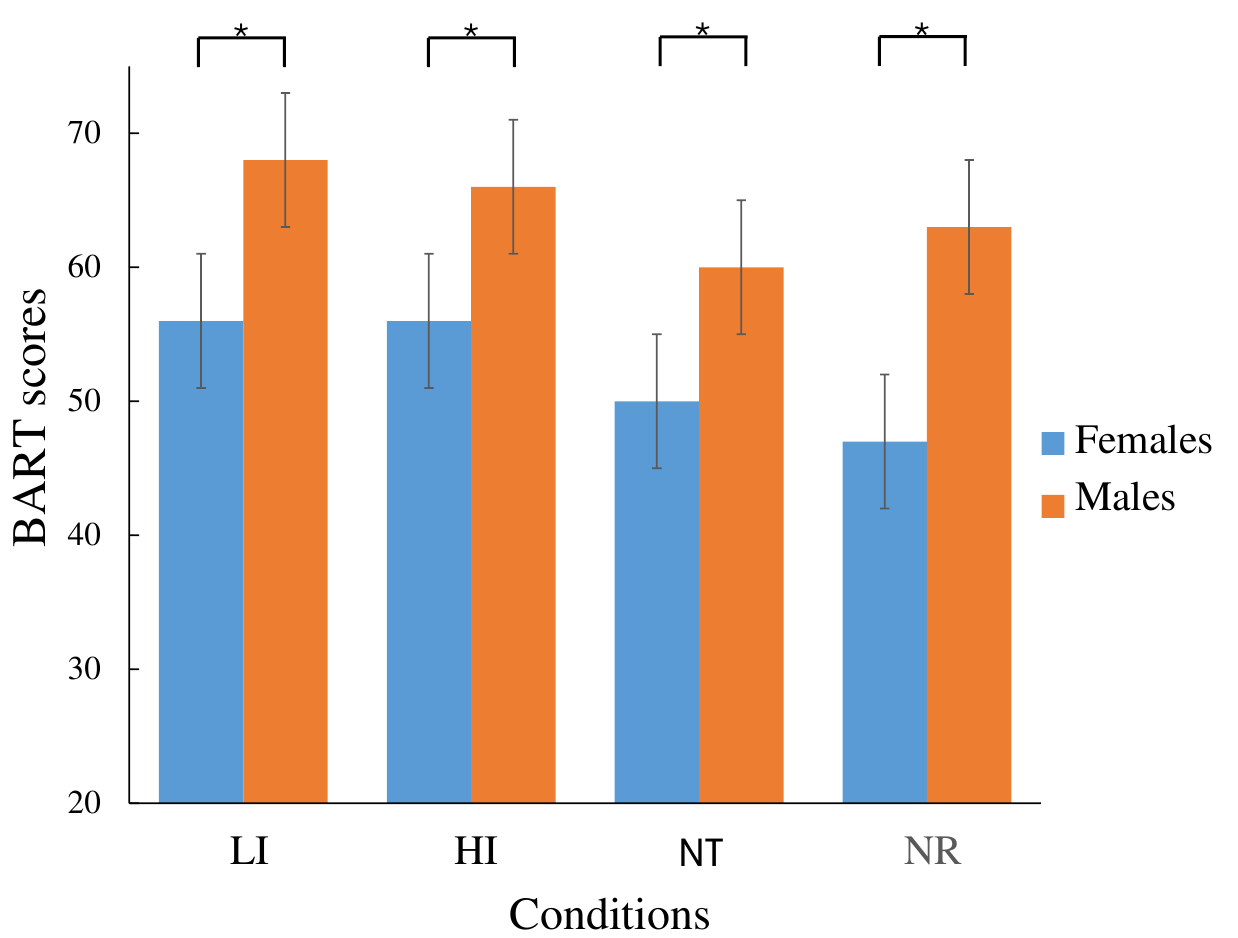}
\caption{\mybluehl{BART scores per gender across conditions. Blue and orange bars show the mean scores for each condition, (LI, HI, NT and NR conditions), with blue for females' BART scores and orange for males' BART scores. Whiskers indicate Standard Error. (* $p < 0.05$).} 
} \label{Gender_bart}
\end{figure}

\begin{figure}
\setlength{\abovecaptionskip}{0.2cm}
\setlength{\belowcaptionskip}{-0.4cm}
\centering
\includegraphics[width=\columnwidth]{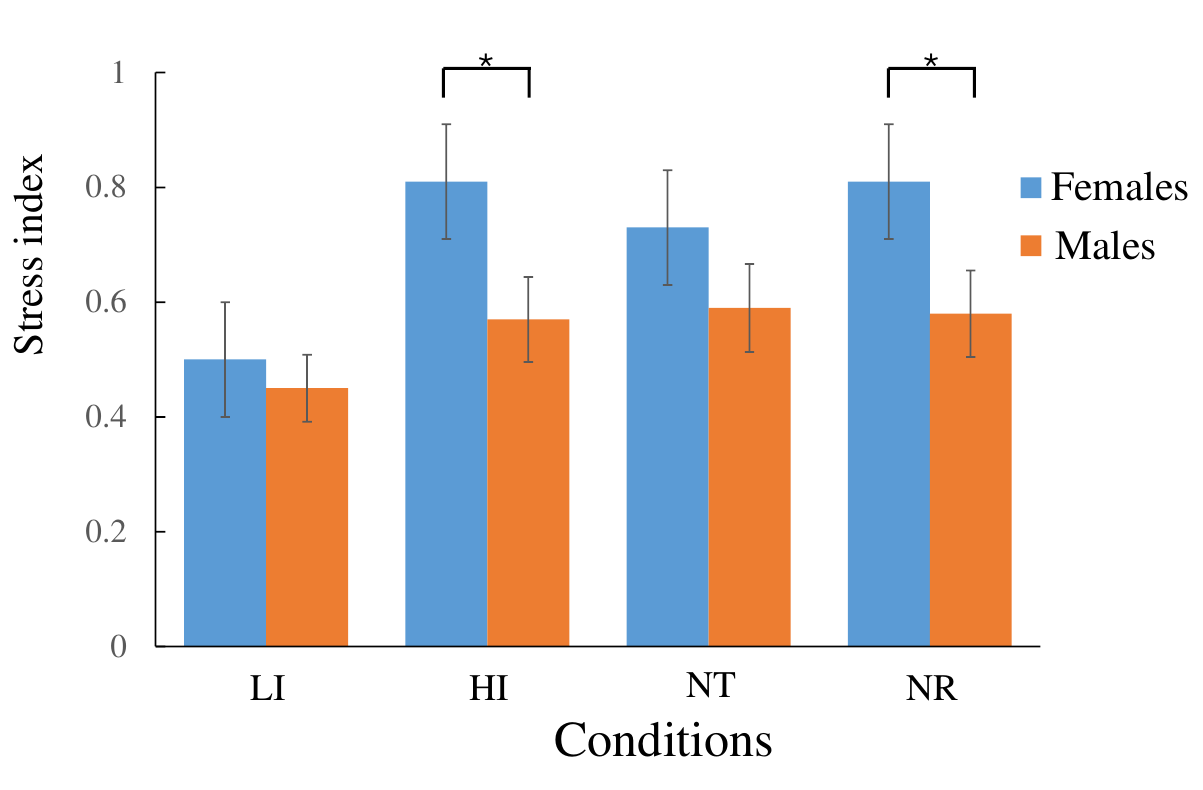}
\caption{\mybluehl{Stress index per gender across conditions. Bars show the average stress index for each condition  (LI, HI, NT and NR conditions), with blue for females' stress index and orange for males' stress index. Whiskers indicate Standard Error. (* $p < 0.05$).} 
} \label{gender_stress}
\end{figure}

A two-way mixed ANOVA is used for different gender analyses and its interaction effects with measurements like BART scores and stress index, which can be seen in Fig.~\ref{F10}, Fig.~\ref{F9}, Fig.~\ref{Gender_bart}, Fig.~\ref{gender_stress}, and Table~\ref{gender}. The analysis of within-subject effects showed no interaction between gender and the different conditions (LI, HI, NT, NR) on BART scores ($F(3, 111) = 1.21, p > 0.05$). \mybluehl{When examining gender as a main effect on BART scores}, we observed a significant main effect of gender on the dependent variable ($F(1, 109) = 23.103, p < 0.001, \eta^2 = 0.391$), which indicates that male ($64.71\pm1.71$) scored significantly higher than the female ($52.47\pm1.71$) during the BART game. Similar results were obtained for the analysis of HRV. There is no significant interaction effect between gender and the different conditions on the stress index; However, when regarding gender as a main effect on stress index, females had a significantly higher stress index than males. ($F(1, 111) = 6.853, p = 0.013, \eta^2 = 0.16$). \mybluehl{In addition, pairwise comparison across four conditions showed that females performed significantly worse than males in BART tasks} in LI $(p < 0.001)$, HI $(p = 0.007)$, NR $(p < 0.001)$ and NT $(p = 0.011)$ conditions, and females experienced significantly higher stress levels than males in HI $(p = 0.011)$ and NR $(p = 0.038)$ conditions. In summary, during risk-taking situations in BART tasks, males exhibited a greater willingness to engage in risky behaviour compared to females and females exhibited lower HRV levels, potentially reflecting increased emotional strain, stress, or anxiety among participants. 


\subsection{Electrodermal activity}
\label{HRIEDA}

\begin{figure}

\centering
\includegraphics[width=\columnwidth]{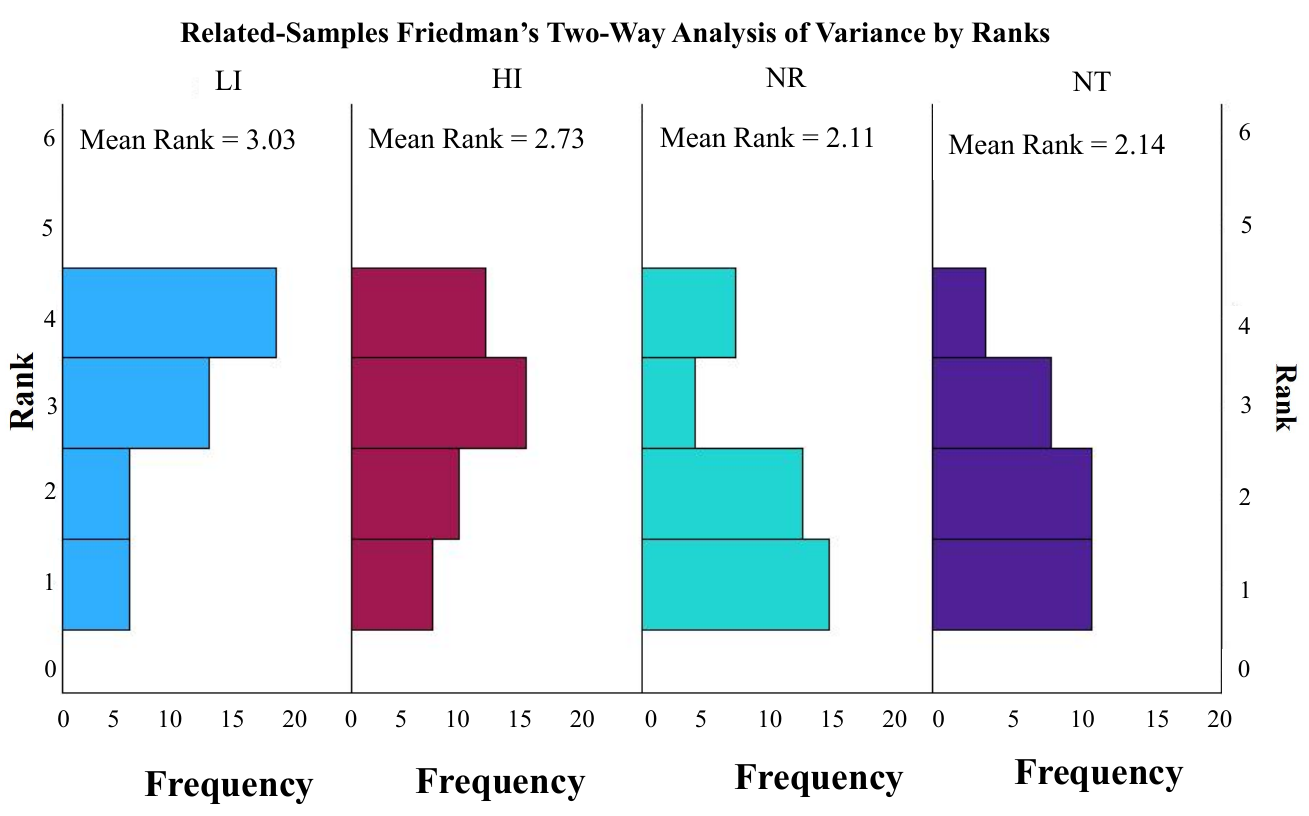}
\caption{\mybluehl{EDA recordings per condition. Bars show the mean ranks of EDA for four distinct conditions labelled LI, HI, NT, and NR, with each condition represented by a different colour. The x-axis denotes the frequency of EDA within each condition, while the y-axis indicates the rank of EDA.} }
\label{EDA_rank}
\end{figure}


To evaluate the EDA across the four conditions, we first normalised the EDA data by subtracting the baseline EDA for each participant to take individual skin conductance differences into consideration. Then, a Shapiro-Wilk test (with $\alpha$ = 0.05) was conducted to assess the data distribution, considering the limited sample size. The test did not provide evidence for a normal distribution of the data. Consequently, a non-parametric test known as Friedman's test (with $\alpha$ = 0.05) was used to compare the previously mentioned EDA data between the conditions. Which was found to be statistically significant and the effect size was measured using Kendall's W, was calculated to be 0.112 across the conditions, suggesting a modest level of agreement ($W(3,112) = 0.112, \chi^2(3) = 13.70,  p=0.003$). 

The mean ranking for LI, HI, NT, and NR conditions is 3.03, 2.73, 2.11, and 2.14 respectively, as shown in Fig.~\ref{EDA_rank}. Further, pairwise comparisons with Bonferroni correction were executed to discern differences between the categories. Notably, the results show that the participants under the LI condition have significantly higher EDA than in NT ($p = 0.003)$ and NR ($p=0.002$) conditions. However, no significant differences were observed between the HI, NT, and NR conditions.

\subsection{Trust}

\begin{figure}[h!]
\setlength{\abovecaptionskip}{0cm}
\setlength{\belowcaptionskip}{0cm}
\centering
\includegraphics[width=\columnwidth]{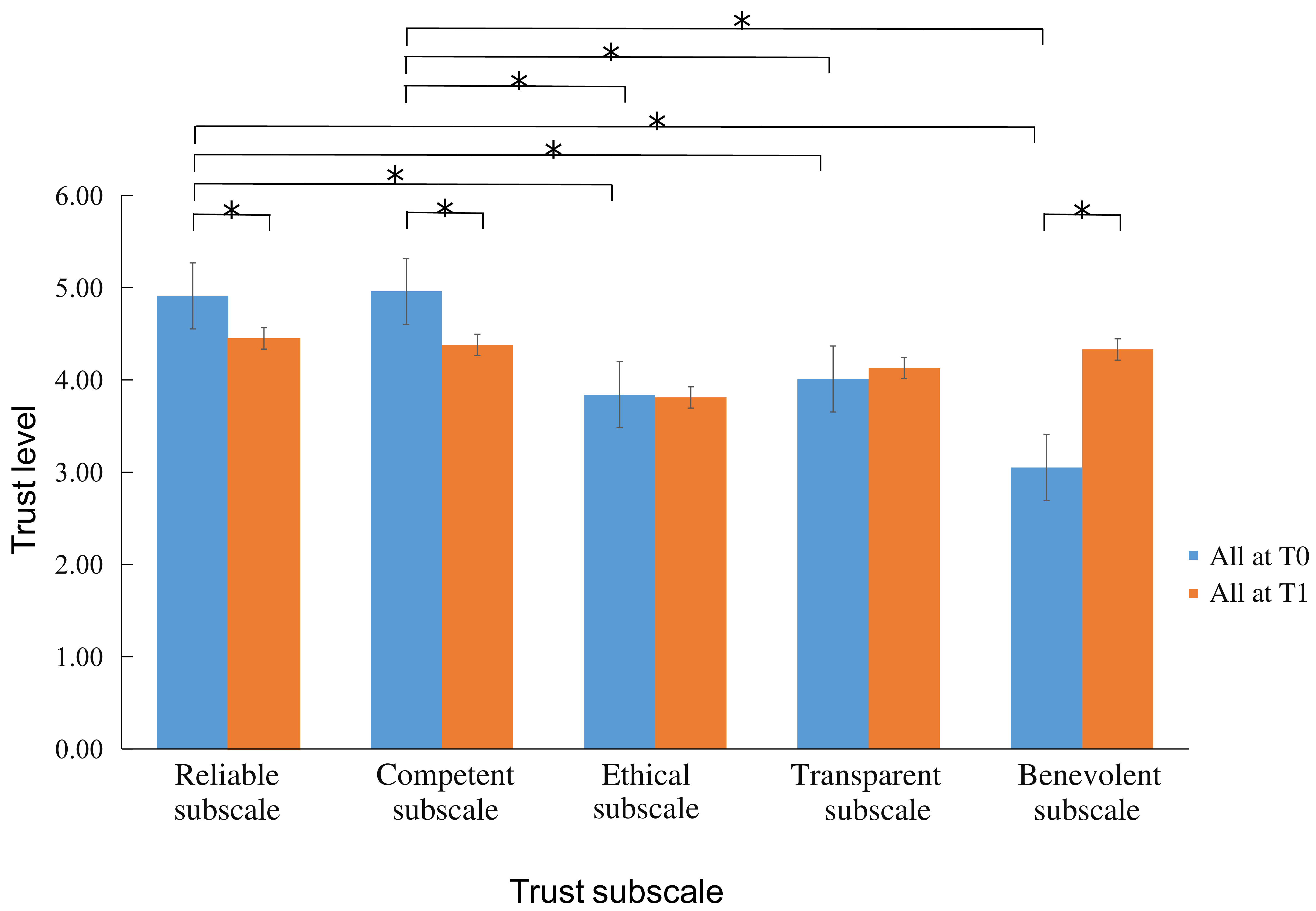}
\caption{\mybluehl{Comparison of trust scores per subscale between time T0 and time T1. Bars show the average trust scores for each subscale at time T0 and time T1, whiskers indicate Standard Error (* $p < 0.05$).}} \label{F8}
\end{figure}

\begin{figure*}[h!]
\begin{minipage}[t]{0.5\textwidth}
\centering
\includegraphics[width=\textwidth]{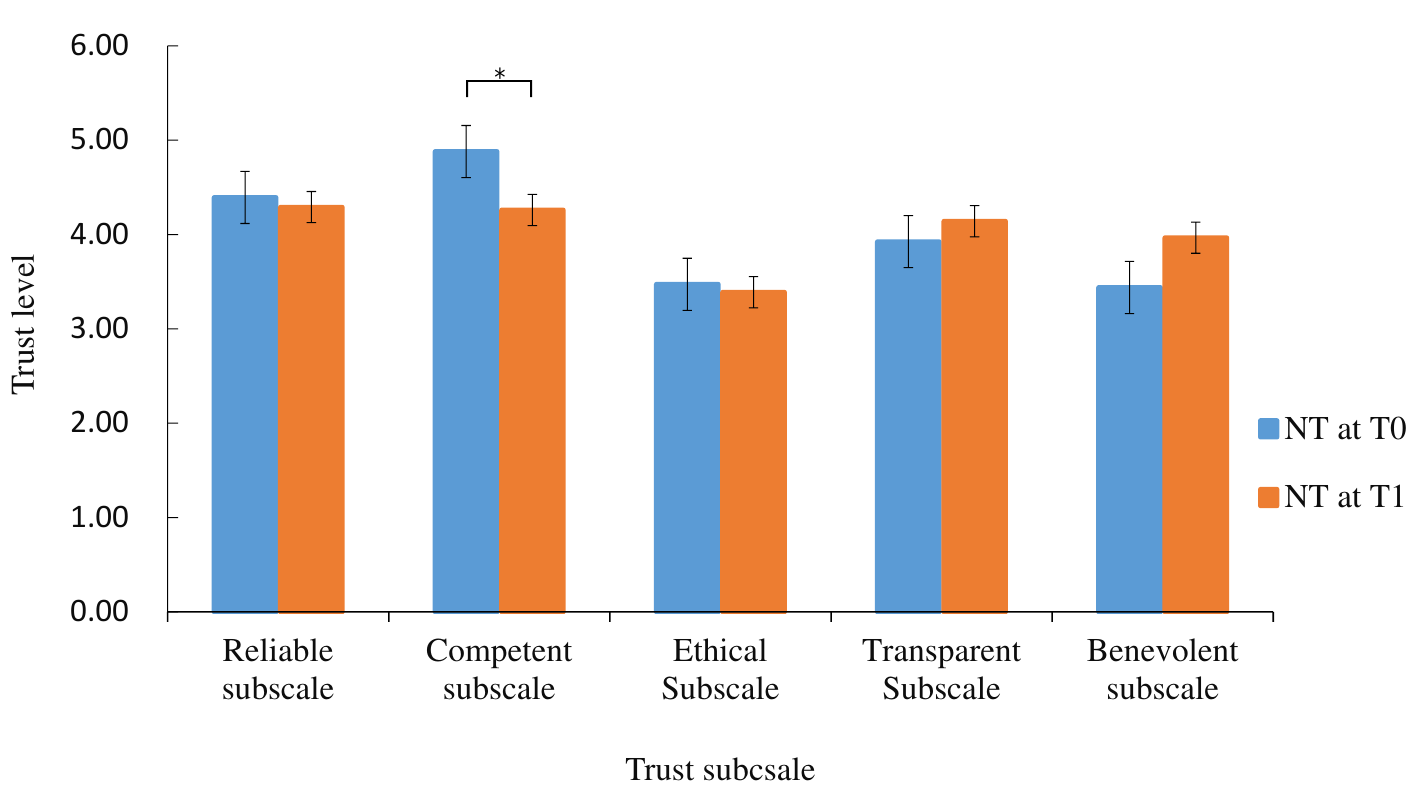}
\caption{Trust measure comparison in the NT condition between T0 and T1, no significant difference was found between T0 and T1.} \label{F4}
\end{minipage}
	\hspace{5mm} 
\begin{minipage}[t]{0.5\textwidth}
\centering
\includegraphics[width=0.9\textwidth]{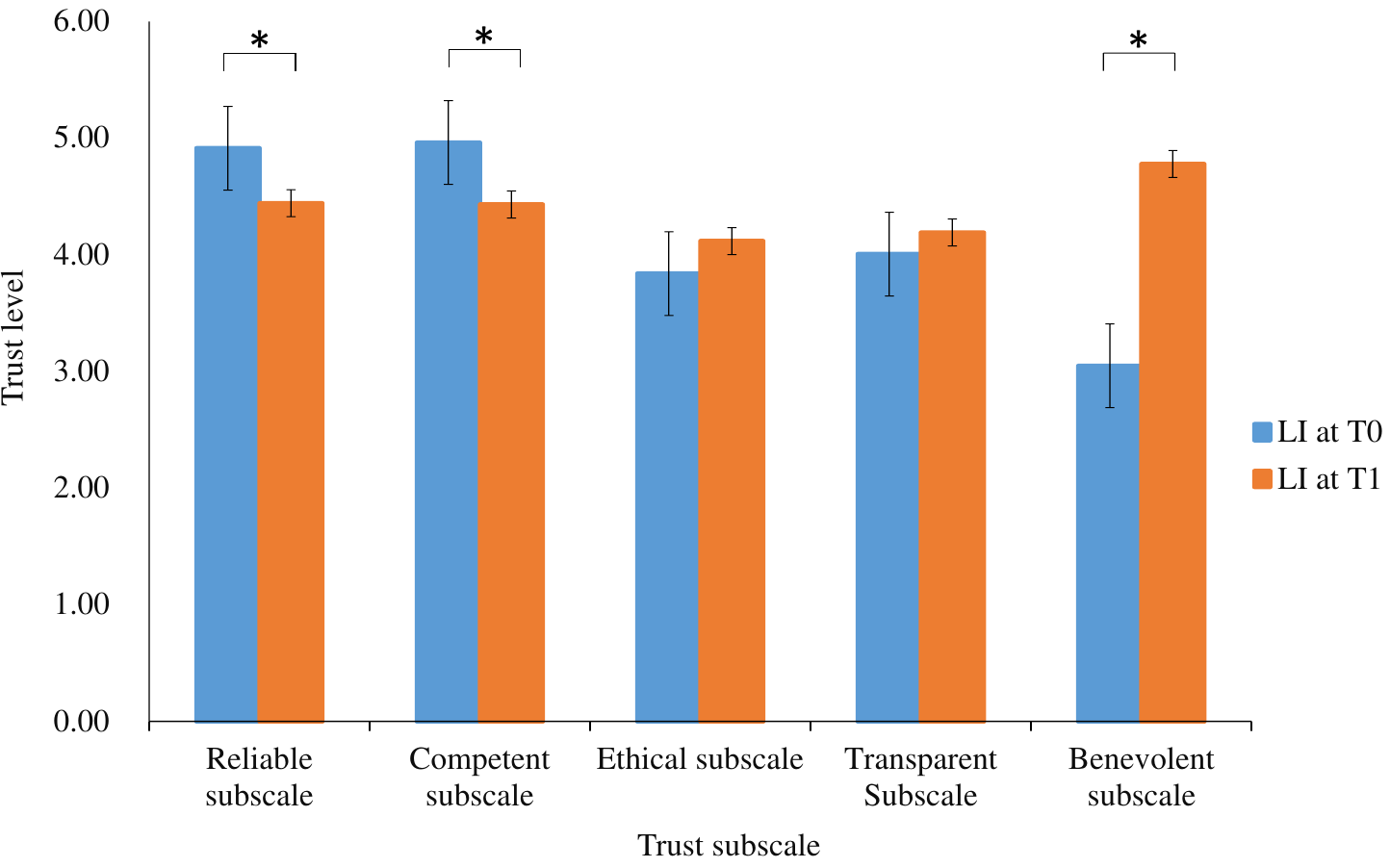}
\caption{Trust measure comparison in the LI condition between T0 and T1, only one significant difference was found between T0 and T1 (* $p < 0.05$).} \label{F5}
\end{minipage}
\end{figure*}

\begin{figure*}[h!]
\begin{minipage}[t]{0.5\textwidth}
\centering
\includegraphics[width=\textwidth]{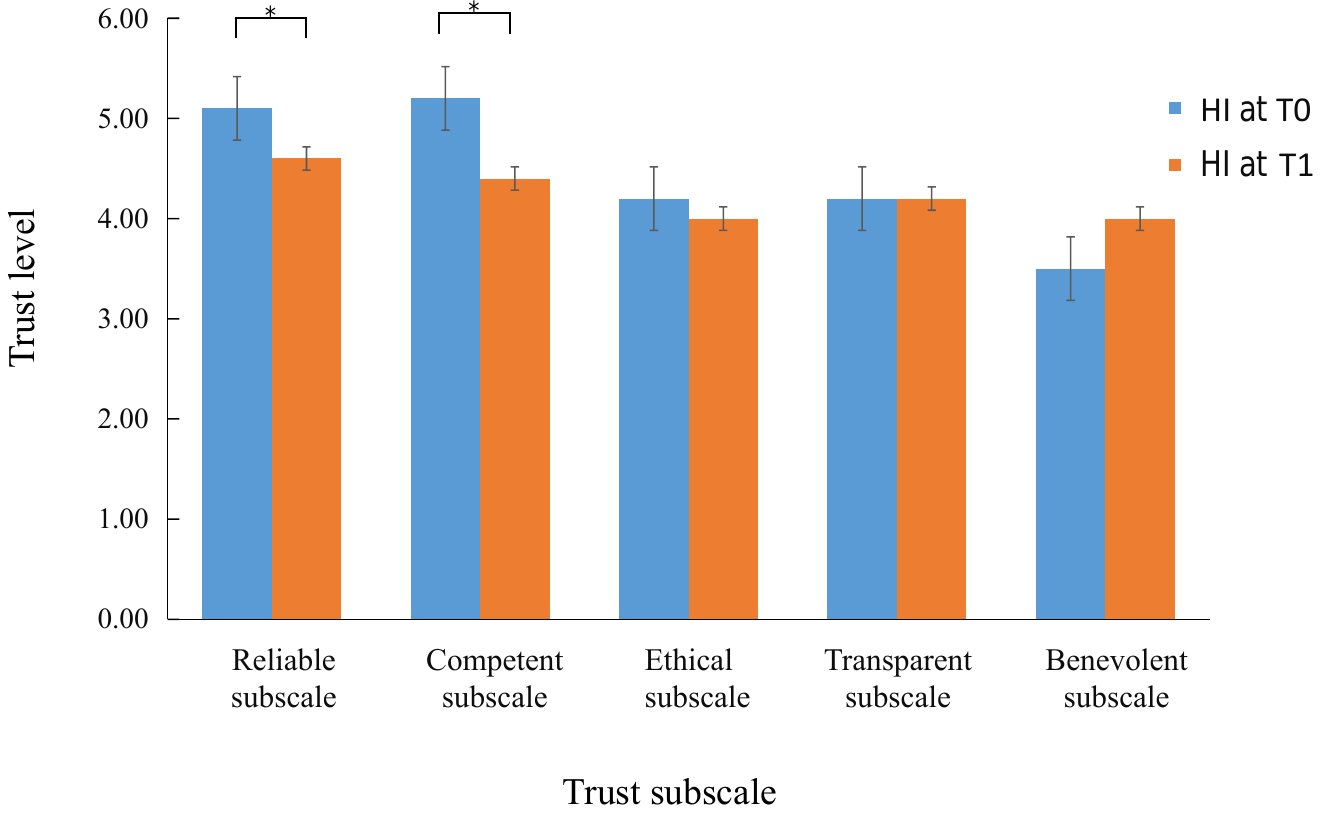}
\caption{Trust measure comparison in the HI condition between T0 and T1, no significant difference was found between T0 and T1.} \label{F6}
\end{minipage}
      \hspace{5mm} 
\begin{minipage}[t]{0.5\textwidth}
\centering
\includegraphics[width=\textwidth]{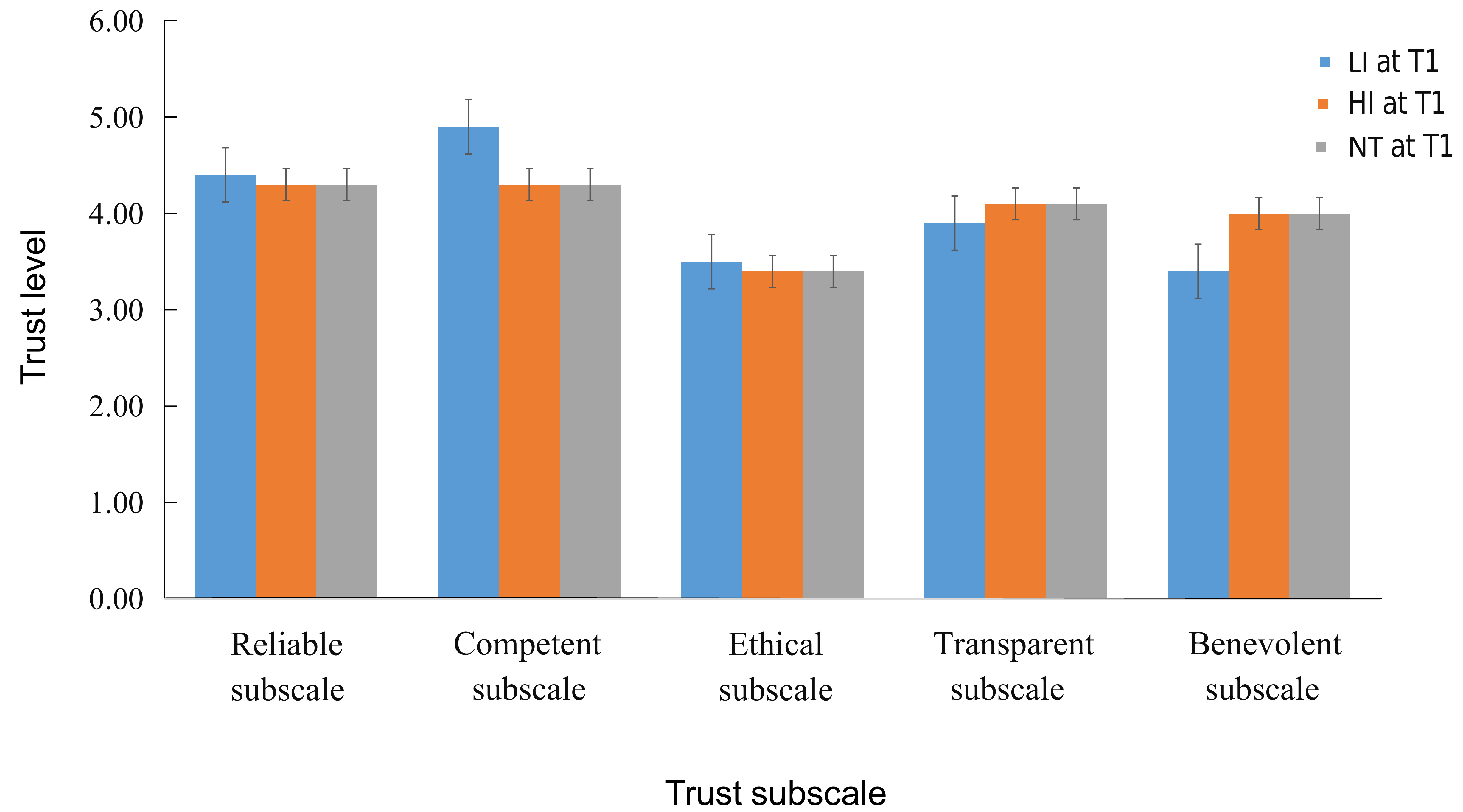}
\caption{\mybluehl{Trust measure comparison for all participants across all conditions at T1, no significant difference was found among all the conditions at T1.}} \label{F7}
\end{minipage}
\end{figure*}

We used MDMT V2 to investigate the effects of human-robot tactile interaction on changes in people's trust, which participants were invited to fill at T0 and T1. MDMT V2 encompasses moral trust and performance trust. Fig.~\ref{F8}, Fig.~\ref{F4}, Fig.~\ref{F5}, and Fig.~\ref{F6} and Table~\ref{tab8} depict the participants' trust scores. The average scores of reliability and competence subscales show a decrease from T0 to T1. Conversely, moral trust ---comprising ethical, transparent, and benevolent subscales--- towards the robot, exhibited an increase during the interaction from T0 to T1. The scores of each subscale of MDMT V2 in the three experimental conditions (HI, LI, and NT) at T0 and T1 are shown in Fig.~\ref{F7} and Table~\ref{tab8}. 

A two-way mixed ANOVA was used on the trust analysis, including five subscales between time T0 and T1 with three experimental conditions. The results revealed no significant differences in any of the subscale measurements among the HI, LI, and NT conditions at T0. Although the LI condition demonstrated slightly higher levels of trust compared to the other conditions, this difference did not reach statistical significance. In addition, Fig.~\ref{F7} shows the trust scores comparison for all participants across all conditions at T1, and no significant difference was found among all the conditions at T1, which indicates that trust levels did not significantly differ among the robot's presence, tactile and verbal interactions at T1. Interestingly, there is a significant main effect on trust scores among the different subscales ($F(4,32) = 4.93, p = 0.003, \eta^2 = 0.24$). The pairwise comparisons with Bonferroni correction show that trust scores on performance trust are significantly higher than moral trust at T0; specifically, the trust scores on the reliable subscale are higher than on the ethical ($p = 0.009$), transparent ($p = 0.002$), and benevolent ($p < 0.001$) subscales. Similarly, the competent subscale scores are higher than the ethical ($p = 0.002$), transparent ($p < 0.001$) and benevolent ($p < 0.001$) subscale scores. This indicates that participants have higher performance trust than moral trust in the robot before the interaction, which can be seen in Fig.~\ref{F8}. However, the post-hoc test result revealed that there is no significant difference among all the subscale trust scores at T1.

Moreover, there is a significant interaction effect between time (T0 and T1) and different trust subscales ($F(4,32) = 8.33, p<0.001, \eta^2 = 0.43$); the pairwise comparisons with Bonferroni correction indicated that a significant increase in trust scores for the benevolent subscale between T0 to T1 ($p<0.001$) in the LI condition, as shown in Fig.~\ref{F5}, whereas a significant decrease in trust scores for the reliable ($p=0.029$) and competent ($p=0.026$) subscales. Similarly, we observed a significant decrease in trust scores for the reliable ($p=0.016$) and competent ($p=0.007$) subscales in the HI condition, as referred to in Fig.~\ref{F6}. In the NT conditions, Fig.~\ref{F4} indicated a significant decrease in trust scores for the competent ($p=0.025$) subscale. As for the overall scores of MDMT V2 (consisting of the subscale scores, performance trust, and moral trust) between T0 and T1, the pairwise comparisons with Bonferroni correction indicated that the trust score on reliable ($p=0.005$) and competent ($p<0.001$) subscales significantly decreased while the trust scores increased on benevolent ($p<0.001$) subscale, which can be observed in the Fig.~\ref{F8}.

\begin{table*}\footnotesize 
\setlength{\abovecaptionskip}{0.0cm}   
	\setlength{\belowcaptionskip}{-0cm}  
	\renewcommand\tabcolsep{2.0pt} 
	\centering
	\caption{Means (and Standard Deviations) of MDMT V2 measurements by Condition (LI, HI, and NT condition).}
	\begin{tabular}
	{
	p{3cm} 
 p{1.6cm}<{\centering}
	 p{1.6cm}<{\centering} 
  p{1.6cm}<{\centering}
	p{1.6cm}<{\centering} 
 p{1.6cm}<{\centering}
	p{1.6cm}<{\centering} 
	} 
\hline
        {\textbf{Variables}} &
        \multicolumn{2}{c}{\textbf{LI}} &
        \multicolumn{2}{c}{\textbf{HI}} &
        \multicolumn{2}{c}{\textbf{NT}}  \\


    \specialrule{0pt}{0pt}{0pt}
          \hline
    \specialrule{0pt}{0pt}{0pt}
      
	Measurements	& \textbf{T0} & \textbf{T1}
  & \textbf{T0} & \textbf{T1}  
  & \textbf{T0} & \textbf{T1} \\

    \hline
     {Reliable subscale }  & $4.9\pm1.6$ & $4.4\pm1.8$ &
      $5.1\pm1.4$ & $4.6\pm1.5$ &
     $4.4\pm1.3$  &  $4.3\pm1.7$  \\

     {Competent subscale }  & $5.0\pm1.6$  & $4.4\pm1.7$  &
    $5.2\pm1.6$ & $4.4\pm1.7$  &
      $4.9\pm1.4$   &  $4.3\pm1.5$  \\

     {Ethical subscaele }  & $3.8\pm2.6$ & $4.1\pm2.2$ &
    $4.2\pm1.8$ & $4.0\pm1.6$ &
     $3.5\pm2.0$  &  $3.4\pm1.8$ \\

     {Transparent subscale}  & $4.0\pm2.3$ & $4.2\pm2.0$ &
     $4.2\pm1.8$ & $4.2\pm1.5$ &
      $3.9\pm1.8$ &  $4.1\pm1.8$ \\
     
    {Benevolent subscale}  & $3.0\pm2.3$ & $4.8\pm1.7$ &
    $3.5\pm1.9$ & $4.0\pm1.8$ &
     $3.4\pm1.9$ &  $4.0\pm1.6$ \\
     
    {Performance trust}  & $4.9\pm1.6$ & $4.4\pm1.8$ &
    $5.2\pm1.5$ & $4.5\pm1.6$ &
     $4.2\pm1.3$  &  $4.3\pm1.6$ \\
     
    {Moral trust}  & $3.7\pm 2.4$ & $4.4\pm2.0$ &
    $4.0\pm1.9$ & $4.0\pm1.6$ &
    $3.6\pm1.8$  &  $3.8\pm1.8$ \\

    {Overall}  & $4.2\pm2.2$ & $4.4\pm1.9$ &
    $4.5\pm1.8$ & $4.2\pm1.6$ &
    $4.0\pm1.8$  &  $4.0\pm1.7$ \\
     \hline

	\end{tabular}
	\label{tab8}
\end{table*}

\subsection{Correlation analysis}

Pearson's correlation coefficient ($r$) was used to assess the relationship between self-reported risk-taking and BART scores, a positive and significant correlation was found between the total BART scores throughout the entire experiment and the self-reported risk-taking scores ($r = 0.43$, $p = 0.007$). This indicates that individuals with a greater inclination towards risk-taking exhibited higher BART scores.

The correlation between the Negative Attitude Toward Robots Scale (NARS) and BART scores is shown in Table~\ref{tab9}. A significant negative correlation was observed between no-touch BART scores and NARS scores ($r = 0.35, p < 0.001$), suggesting that participants with a negative attitude towards the robot exhibit lower risk-taking tendencies. No significant correlation exists between no-robot BART scores and NARS scores, which makes sense as no robot was present to influence risk-taking behaviour. 

Interestingly, as shown in Table~\ref{tab9}, the correlation becomes more pronounced in conditions where participants have tactile interaction with the robot, as there is a stronger negative correlation between BART scores and NARS scores.  Moving to the overall BART scores and NARS correlations, there is a negative correlation between the overall BART scores and NARS scores ($r = -0.57$, $p < 0.001$). Notably, a negative correlation was found between NARS scores and trust ($r = -0.38, p = 0.018$), indicating that a more negative attitude towards the robot translates to lower trust in robots. However, no significant difference was found between trust and physiological measurements, specifically EDA (Electrodermal Activity) and HRV (heart-rate variability).




\begin{table*}[t]\footnotesize 
\setlength{\abovecaptionskip}{0cm}   
	\setlength{\belowcaptionskip}{-0cm}  
	\renewcommand\tabcolsep{2.0pt} 
	\centering
	\caption{Correlations between NARS scores and BART scores.  ** at $p < 0.001$.}

	\begin{tabular}
	{p{3cm} p{2cm} p{2cm} p{2cm} p{2cm}	} 
\hline
    
    {\textbf{BART performance}}& 
     {\textbf{HI scores}}  & 
     {\textbf{LI scores}}  & 
     {\textbf{NT scores}}  & 
     {\textbf{NR scores}} \\
\hline 
{Negative attitudes to\-wards robots (NARS)} & 
  \multirow{2}{*}{-0.49**} &
   \multirow{2}{*}{-0.64**} &
  \multirow{2}{*}{-0.34*} &
 \multirow{2}{*}{-0.19,  $p > 0.05$} \\

     \hline
	\end{tabular}
	\label{tab9}
\end{table*}

\section{Human-object interaction experiment results and analysis}
\label{sec4:hoi_ex}

\subsection{BART performance}
\begin{figure}[h!]

\centering
\includegraphics[width=\columnwidth]{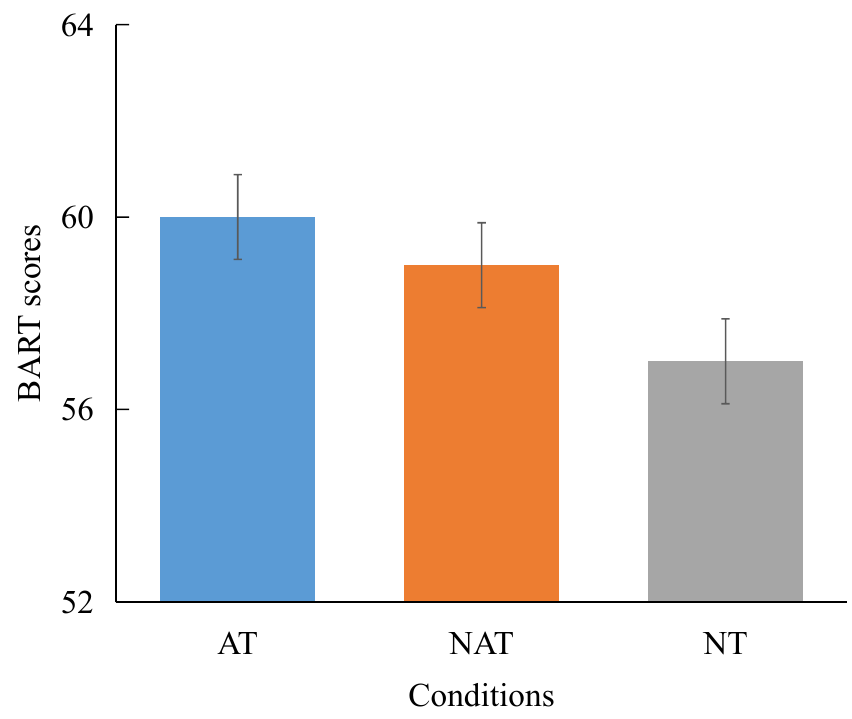}
\caption{BART scores per condition. Bars show the average scores for each condition (LI, HI, NT and NR conditons), whiskers indicate Standard Error. (* $p < 0.05$).} \label{F18}
\end{figure}

A one-way repeated measures analysis of variance (ANOVA) was conducted to examine the differences in BART scores across the three interaction conditions. The results showed that there were no significant differences in BART scores ($F(3, 111) = 1.18$, $p > 0.05$) between the no tactile interaction condition ($57 \pm 2$), AT condition ($60 \pm 2$) and NAT condition ($59 \pm 2$). These findings indicate that neither the AT nor NAT conditions were effective in promoting participants to engage in more risk-taking behaviour (see Fig.~\ref{F18}).

\subsection{Heart-rate variability}

The results of the SNS stress index and PNS index can be found in Table~\ref{tab10}. The stress index and SNS index were the lowest in the NT condition, while the PNS index was the highest. This indicates that participants experienced the lowest stress level in the NT condition.

To further analyse these results, A one-way repeated measures ANOVA and post hoc tests were conducted. However, the results revealed no significant difference between the three conditions, indicating that tactile interaction with the box had no impact on reducing stress.

\begin{table*}[h!]\footnotesize 
\setlength{\abovecaptionskip}{0cm}   
	\setlength{\belowcaptionskip}{-0cm}  
	\renewcommand\tabcolsep{2.0pt} 
  \caption{Human-object interaction HRV analysis}
  \centering
  \label{tab10}
  \begin{tabular}{p{2.5cm}p{2cm}p{2cm}p{2cm}p{2cm}}
    \hline
     \textbf{Conditions} & \textbf{AT} &  \textbf{NAT} & \textbf{NT} & \textbf{F(2,105)}\\
    \hline
    PNS index & $51.5\pm6.8$ & $50.0\pm7.6$ & $57.2\pm8.5$ & $1.59, p>0.05$\\
    SNS index & $-2.3\pm0.2$ & $-2.2\pm0.18$ & $-2.5\pm0.2$ &  $1.97, p>0.05$\\
    Stress index & $0.92\pm0.07$ & $0.87\pm0.07$ & $0.85\pm0.06$ & $0.87, p>0.05$ \\
    \hline
\end{tabular}
\end{table*}

\subsection{Electrodermal activity}
\label{sec5:discussion}

We followed the same data analysis as in section \ref{HRIEDA}. We recorded data when the participants touched the object and compared it with the time of EDA peaks in the low-intensity robot touch condition. For the AT and NAT conditions, we respectively see 31\% and 26 \% coincidence. These findings suggest that there were no significant EDA peaks observed when participants engaged in tactile interaction with the box. Given that the EDA data did not conform to a normal distribution, a non-parametric analysis, namely Kendall's Coefficient of Concordance (with $\alpha = 0.05$), was used.  The post-hoc analysis revealed no significant differences among the three conditions. These results indicate that participants did not exhibit an emotional response during their tactile interactions with the box.

\section{Discussion}
\label{sec:discussion}

\subsection{Results and interpretation}


\subsubsection{Risk-taking behaviour}

Earlier studies have shown that verbal encouragement by a robot can encourage individuals to take more risk \cite{hri2}. Surprisingly, our study failed to replicate this finding. However, our research offers new perspectives by revealing that tactile interaction with a robot can increase risk-taking behaviour while at the same time reducing the stress experienced during risk-taking. Both the low-intensity, gentle touching of the robot and high-intensity tactile contact (where participants held the robot on their lap) were found to promote risk-taking behaviour. In contrast, interaction without touching the robot did not yield any significant effect on participants' inclination towards risk-taking behaviour. These findings highlight the importance of tactile interaction as a factor in influencing risk-taking tendencies.

\subsubsection{Physiological responses}


In our research, we observed a significant increase in the average level of EDA when people gently touch the robot, compared to intense tactile interaction with the robot or no tactile interaction at all. In addition, when participants had physical contact with the robot during gentle tactile interactions distinct peaks in SCR were observed, indicating a strong emotional response towards the robot. In contrast, the EDA levels do not differ when people are having an intense physical interaction ---by taking the robot on their lap, when they are not touching the robot at all or when no robot is present. This implies that the emotional states of participants were relatively consistent across these conditions. Surprisingly, only gentle interaction with the robot elicits an emotional response. The specific nature of this emotional response is yet to be determined, but it could be posited that the low-intensity interaction captured the attention of participants, and the interaction behaviour was more engaging than in other conditions, which may have resulted in a change in the EDA level. One possible explanation for these findings object and compared it with the time of EDA peaks in the low-intensity robot touch condition. These results indicate that participants did not exhibit an emotional response during their tactile interactions with the box. These findings suggest two key insights: (1) Touching the robot elicits EDA peaks, indicating an emotional response in individuals. (2) Participants exhibit higher skin conductivity levels during robot interactions compared to other interaction periods.

Our study revealed significant differences in stress levels based on physiological indicators of parasympathetic and sympathetic nervous system activity, as well as a Baevsky’s stress index (stress index), during low-intensity and high-intensity tactile interaction. The results indicate that participants experienced lower stress levels during low-intensity tactile interaction, while higher stress levels were observed when the robot was present but not being touched or when no robot was present. These findings demonstrate the impact of tactile interaction on stress regulation, and suggest that low-intensity tactile interaction may be a more effective stress-reducing strategy compared to high-intensity interaction. Additionally, the majority of participants (90\%) reported that the high-intensity condition of holding the robot on their lap felt similar to holding a baby, which may have contributed to feelings of responsibility and stress. In addition, the weight of the robot resting on the participant's lap and the moderate heat generated by the robot are likely to be contributing factors. In contrast, the participants reported experiencing the robot as comforting and encouraging during low-intensity touch interaction.

In conclusion, tactile interaction with a social robot
influence people’s physiological responses.

\subsubsection{Social nature}
To study if the social character of the robot was implicated in these results, we set up a follow-up study in which people interacted with a box --- an obviously non-social object. This revealed that there were no significant differences: whether or not participants interacted with the box through touch or not, the EDA, heart rate variability (HRV), and risk-taking (as measures by the BART performance) did not change. 
This interaction differs markedly from the one with a robot: the robot provides affective gestures, like shaking hands or high five, while the box does not.

In summary, tactile interaction with the box did not have a significant impact on participants' risk-taking behaviour, stress levels or emotional responses. In our first human-robot interaction study, the robot recognised the explosion of the balloon and provided comfort, encouraged participants to continue pumping the balloon, and engaged in touch-based interactions. The social nature of the tactile interaction is related to (1) the Nao's physical presence and social appearance, and (2) the conveyed emotion during the interaction. When comparing the results from Study 2 (human-object interaction) with the results from the interaction with a social robot, it is clear that tactile interaction with the robot is a social activity with social consequences, rather than simply a physical event.

\subsubsection{Trust}

Trust plays a crucial role in fostering societal well-being as it represents social capital and has a positive impact on social well-being. As the interaction with the robot progressed, we observed a decrease in performance trust levels, while moral trust levels increased. However, there was no significant difference in performance trust and moral trust levels between the initial measurement (T0) and the final measurement (T1). 
One possible explanation for this is that the robot encouraged participants to pump more during the BART task, ignoring the risks involved. This encouragement led to a decrease in performance trust in the robot, as participants may have been concerned about the possibility of the balloon bursting. At the same time, the comforting and positive tactile interaction with the robot helped to form a positive perception of the robot, as they ranked higher on the robot's positive emotional state during the tactile interaction than their negative emotional state. Previous research has shown that the performance and characteristics of robots have the greatest association with trust \cite{hancock2011meta}. The affective and supportive interaction may have helped to increase participants' moral trust in the robot. However, the incitement to take risks may have led to a negative impression of the robot, resulting in a decrease in moral trust. Previous research has suggested that errors, task type, and personality influence human-robot cooperation and trust \cite{salem2015would}. Our results shown that tactile interaction influences trust in the social robot.

\subsubsection{Gender effects}

In our study, we observed that male participants demonstrated a higher propensity for risk-taking and exhibited lower stress levels compared to female participants, a finding that aligns with the literature on gender differences in risk perception \cite{byrnes1999gender} and stress response. Significantly higher scores in risk-taking were recorded for males across all experimental conditions.

Concerning stress levels, our data revealed that female participants reported higher stress levels than male participants. This difference was statistically significant in the HI and NR conditions, however, participants in the LI condition did not yield a significant difference in stress levels between genders, which may indicate that low-intensity tactile interaction could narrow the stress response gap. Similarly, the NT condition appeared to slightly reduce this gap, though not significantly. In the HI condition, the requirement for continuous physical contact with a robot exacerbated the stress levels gap, with the female response being stronger than the male response. One possible explanation for these findings is the influence of tactile or affective tactile interactions on stress levels, as previous research has indicated that females tend to have stronger emotional responses to affective touch compared to males \cite{chan2020effect}. In summary, tactile interaction influences participants' risk-taking behaviour and stress levels differently across genders.


\subsection{Limitations}

The study has several constraints that could affect the generalisability of the results. The demographic composition of the sample, encompassing cultural background, age, and personality differences, along with the participant's proximity to the robot, the robot's design, and the nature of the interaction ---whether it involves active or passive touch--- may all influence the outcomes. These factors should be considered when drawing conclusions from the results. In addition, participant engagement might have been further enhanced by offering real monetary rewards in the BART task.

\subsubsection{Participant population}

First, the population was limited in diversity, potentially impinging upon the ability to extrapolate the study's outcomes to broader contexts. Preferences regarding interpersonal distance, body orientation, and touch in dyadic interactions appear to be influenced by an interactant's culture, gender, and age \cite{remland1995interpersonal}. Given the lack of diversity in our participant demographic, we acknowledge that our results might not transfer to demographics different from ours.


Hall \cite{hall1966hidden} argued that people from ``contact” cultures prefer more tactile interacton than those from ``noncontact” cultures. This implies that interactants from contact cultures (e.g., Arabs, Latin Americans, southern Europeans), when compared with those from noncontact cultures (e.g., Asians, North Americans, northern Europeans) tend to interact at closer distances, maintain more direct body orientations, and touch more frequently. In our study, all participants were recruited in Belgium and were from both Northern Europe and Asia cultures, as such we assume that the majority of our participants were from a low contact culture.

Individual differences in the use of tactile communication can be influenced by various personality traits and characteristics \cite{deethardt1983tactile, reniers2016risk}. In this research, we didn't take this into consideration, which could also influence how we interpret our results.

\subsubsection{Distance towards the robot}

Previous research, notably Hall's foundational studies, has demarcated specific zones of interpersonal space corresponding to varying levels of intimacy within a culture. These distances are categorised as public, social, personal, and intimate distances \cite{velloso2015feet, hall1982hidden}. In the realm of human–robot interaction, studies suggest that individuals may transfer human-to-human social distance norms, to their interactions with robots \cite{kim2014social}. However, in our study, we did not consider human-robot social distance. Consequently, our findings might not hold true in situations where robots infringe upon these established interpersonal spaces. A plausible reason for heightened pressure observed among participants during high-intensity interactions could be the robot's encroachment into the intimate zone, a space typically reserved for close personal relations. It is worth speculating that individuals might be uncomfortable having a robot in close proximity. Similarly, varying the robot's distance, either too close or too far, might yield different interaction outcomes. We recommend future research to factor in these spatial dynamics to gain a comprehensive understanding of human-robot tactile interactions.

\subsubsection{Robot dimensions}

The use of the Nao robot as a research tool also presents some limitations. While the robot’s ``cute'' factor may be appealing to some participants, its capabilities are limited and may not be sufficient for more sophisticated research on human–robot interaction.  Our findings with the Nao robot may not be universally applicable to all types of robots, and our findings might not transfer to, for example, taller robots.

\subsubsection{Active and passive tactile interaction}

One of the objectives of our research was to explore the emotional and stress-related responses of individuals during tactile interactions with robots, however, only robot-initiated interaction was used in this study --- participants were invited to touch the robot, and no participant declined to touch the robot. We observed positive outcomes in low-intensity (LI) conditions, such as reduced stress levels. Similar results were also reported in \cite{geva2020touching, itoh2006development} with passive interaction. However, some mirroring patterns were found in certain human-robot interaction (HRI) studies. The existing literature paints a mixed picture, some studies report positive outcomes for active tactile interactions, but not for passive interactions\cite{chen2011touched, nakagawa2011effect}. In addition, Block \emph{et al.} found that the robot’s responses and proactive gestures were positively perceived by users \cite{block2023arms}. Others reported negative consequences of active tactile interactions from a robot, particularly concerning the reliability of robots that initiate contact \cite{bock2018your}. Moreover, some scholars have highlighted the importance of carefully integrating autonomous and social behaviours, such as communicative touch, in interactions with social agents. They suggest that the specific combination of these behaviours can critically influence whether the impact on attitudes towards embodied social agents is positive or negative \cite{cramer2009give}.

Determining the exact reasons for the positive effects of a robot's touch is complex, especially when contrasting our study with those yielding contrary results. Various factors contribute to this complexity: the robot’s appearance, the texture and feel of its touch, personality traits, assigned tasks, cultural contexts, the attraction caused by the physical interaction and distinctive touch-related behaviours. Each of these aspects can shift the user's perception of the robot.

\subsubsection{Human-robot interaction scenarios}

Ideally, social robots should be evaluated in real-world environments using objective and analytical methods. However, most social robots ---including ours--- are typically tested in lab settings, which may not reflect actual social interactions. The presence of a robot, the context of the study, and the laboratory setting itself could affect how participants interacted with the robot  \cite{irfan2018social}. A true assessment of a robot's social skills requires testing in varied, real-world social environments where it operates independently \cite{sabanovic2006robots}. In addition, the second study only mirrors the settings in the LI condition and the NT condition from the first study, lacking the HI condition, which is putting the box on the participant's lap.

Even though the BART is a reliable and validated tool for assessing real-world risk-taking propensity and is commonly used in research \cite{lejuez2002evaluation}, people's risk-taking behaviour and arousal level can also be influenced by factors over which we have no control outside the lab, such as people's mood or context. While lab studies are foundational and provide important insights, further work in naturalistic settings would be valuable.

\subsubsection{Persuasion and risk-taking behaviour}

Previous research explored older siblings' persuasive influence on young children's risk-taking decisions, which showed that younger children significantly changed their decisions from initially more risky decisions to less risky decisions and vice versa \cite{morrongiello1997sibling}. In our study, we explored whether participants could be influenced by a robot to engage in greater risk-taking actions through both verbal encouragement and tactile interaction, to examine the potential effects of peer pressure on their willingness to take risks. While our experiment did not examine if the robot's persuasion of less risk-taking behaviour could lead to more cautious behaviour, it presents an intriguing possibility for future research—to investigate whether a robot advising against risk could reduce risk-taking tendencies. The interplay between persuasion, tactile engagement, and risk propensity could be explored more in the future; tactile interactions may not uniformly encourage risk-taking. It is conceivable that the combination of persuasive dialogue and tactile feedback could synergistically impact the participants' risk-taking behaviours. In addition, researchers explored different social roles' impact on children's risk-taking decisions; the results showed that half of the children might be persuaded by their best friend and half by their parents \cite{morrongiello2004identifying}, which may indicate that the trust or social bonds between the robot and participants may also influence the persuasion results, lead to different risk-taking behaviours.

\subsection{Conclusion}
Our study investigated the impact of tactile interaction with a social robot on risk-taking behaviour and stress levels. We show that both low-intensity and high-intensity tactile interaction encourages people to take more risks. However low-intensity touch elicited a stronger emotional response and reduced stress levels. 
We also observed a decrease in performance trust together with an increase in moral trust across all experimental conditions as the interaction unfolds. The intention and characteristics of the robot are important factors influencing trust levels, and future research should explore the influence of different robot characteristics on trust.

The study also examined gender effects in performance and psychosocial measurements during the interaction, revealing that male participants showed a greater inclination towards risk-taking and displayed lower levels of stress compared to female participants, confirming the gender effects in risk-taking behaviour reported in the literature \cite{byrnes1999gender}. Notably, during low-intensity tactile interaction, female participants exhibited slightly lower stress levels than males. Under other conditions, female participants experienced higher stress levels.


The study found that self-reported risk-taking behaviour correlated well with the risk they took in a task, as might be expected. Additionally, participants who held a more negative attitude towards robots took less risk, even when the robot encouraged them to take risks. 
However, when invited to touch the robot that would change and they would take more risk when encouraged to do so by the robot. This suggests that tactile interaction mitigated possible negative preconceptions about robots. This ties in with Block \emph{et al.}, who found that users experienced the robot as more natural, enjoyable, and intelligent after tactile interaction \cite{block2023arms}. Moreover, our study revealed a significant positive correlation between risk-taking and the participants' trust in the robot. 
This finding aligns with previous research indicating that initial impressions of a robot can influence human-robot trust in problem-solving scenarios \cite{xu2018impact}. Similarly, the study suggests that the initial impression of the robot impacts trust levels, which in turn influences risk-taking behaviour.

An important result is that the strong affective response of people, as measured by their physiological response, is not caused by the physical action of touching the robot, but instead by people perceiving the robot as a social entity. Touching a non-social object, such as the plastic box in our studies, does not elicit an affective response. This shows that the robot is more than just an object to us, instead, the robot has a social presence and touching a robot has special significance.

In conclusion, social robots have a role to play in promoting physical and emotional well-being and this can be enhanced by designing them to allow tactile engagement \cite{van2013touch}. The results of this study indicate that tactile interaction with a social robot can influence people's physiological responses and decisions and that these are modulated by people's attitudes towards the robot. Overall, the use of tactile interaction in Human-Robot Interaction has the potential to improve the physical and emotional well-being of individuals, particularly in the context of assistive and companion robots. 

\backmatter

\bmhead{Data availability}

Data can be made available for research purposes upon reasonable request to the corresponding author.

\bmhead{Acknowledgments}

Qiaoqiao Ren is partially supported by the China Scholarship Council (CSC). This research is partially funded by the Flemish Government under the ``Onderzoeksprogramma Artifici{ë}le Intelligentie (AI) Vlaanderen'' programme and the Horizon Europe VALAWAI project (grant agreement number 101070930). We thank Yaniv Hanoch for his kind advice on the BART protocol.

\newpage











\vfill\pagebreak\label{sec:refs}
\bibliographystyle{sn-mathphys}
\bibliography{mybib}

\end{document}